%% file: 005_paper_revision_STAR.tex
\documentclass[times, review, 10pt]{elsarticle}

\usepackage{amssymb}

\usepackage{lineno}
\usepackage{multirow}
\usepackage{bbm} 
\usepackage{soul}
\usepackage{amsmath}
\usepackage{color}
\usepackage{hyperref}

\usepackage{amsmath}
\usepackage{amssymb}
\usepackage{amsfonts}
\usepackage{booktabs, makecell, multirow, tabularx}
\usepackage{textcomp}
\usepackage{stfloats}
\usepackage{url}
\usepackage{verbatim}
\usepackage{graphicx}
\usepackage{xcolor}
\usepackage{spacingtricks}
\usepackage{bbm}
\usepackage{multirow}
\usepackage{makecell}
\usepackage{graphicx}
\usepackage{amsmath}
\usepackage{booktabs}
\usepackage{multirow}
\usepackage{array}
\usepackage{amsfonts}
\usepackage{bbm} 
\usepackage{xcolor}
\usepackage{mdframed}
\usepackage{makecell}
\usepackage[normalem]{ulem}
\usepackage{graphicx}
\usepackage{subcaption}
\usepackage{amsmath,amsfonts,bm}

\usepackage{setspace}

\usepackage{algorithmic}
\usepackage{algorithm}

\journal{Pattern Recognition}

\begin{document}




\label{sec:Titlepage}
\input{Titlepage}


\section{Introduction}
\label{sec:Introduction} 

Face anti-spoofing (FAS) aims to distinguish spoof faces from live ones to enhance the security of facial identification systems. 
Most existing FAS methods \cite{huang2025channel,huang2023ldcformer,wang2023domain} adopt a two-class classification strategy that learns discriminative representations from both live and spoof samples. 
While two-class FAS methods perform well when test attacks are covered by the training data, they often fail to generalize to out-of-distribution (OOD) spoof attacks during deployment. 
In contrast, one-class FAS approaches \cite{baweja2020anomaly,huang2021one,lim2020one}, which learn liveness representations solely from live samples, exhibit stronger robustness to unseen spoof types. 
However, this advantage comes at the cost of inferior discriminability and limited cross-domain generalizability. 
Specifically, learning from live-only data causes liveness feature representations to be entangled with domain-related factors, making one-class FAS models sensitive to domain shifts and less competitive than two-class counterparts. 

To address these limitations, we propose a Self-supervised Feature Disentanglement and Augmentation Network (ssFDANet) for one-class FAS. The core idea of ssFDANet is to enrich one-class training with disentangled and augmented liveness and domain features derived solely from live images, thereby bridging the performance gap between one-class and two-class FAS methods without requiring spoof samples or domain annotations.

\begin{figure*}[t] 
    \centering 
     \includegraphics[width=0.98\textwidth]{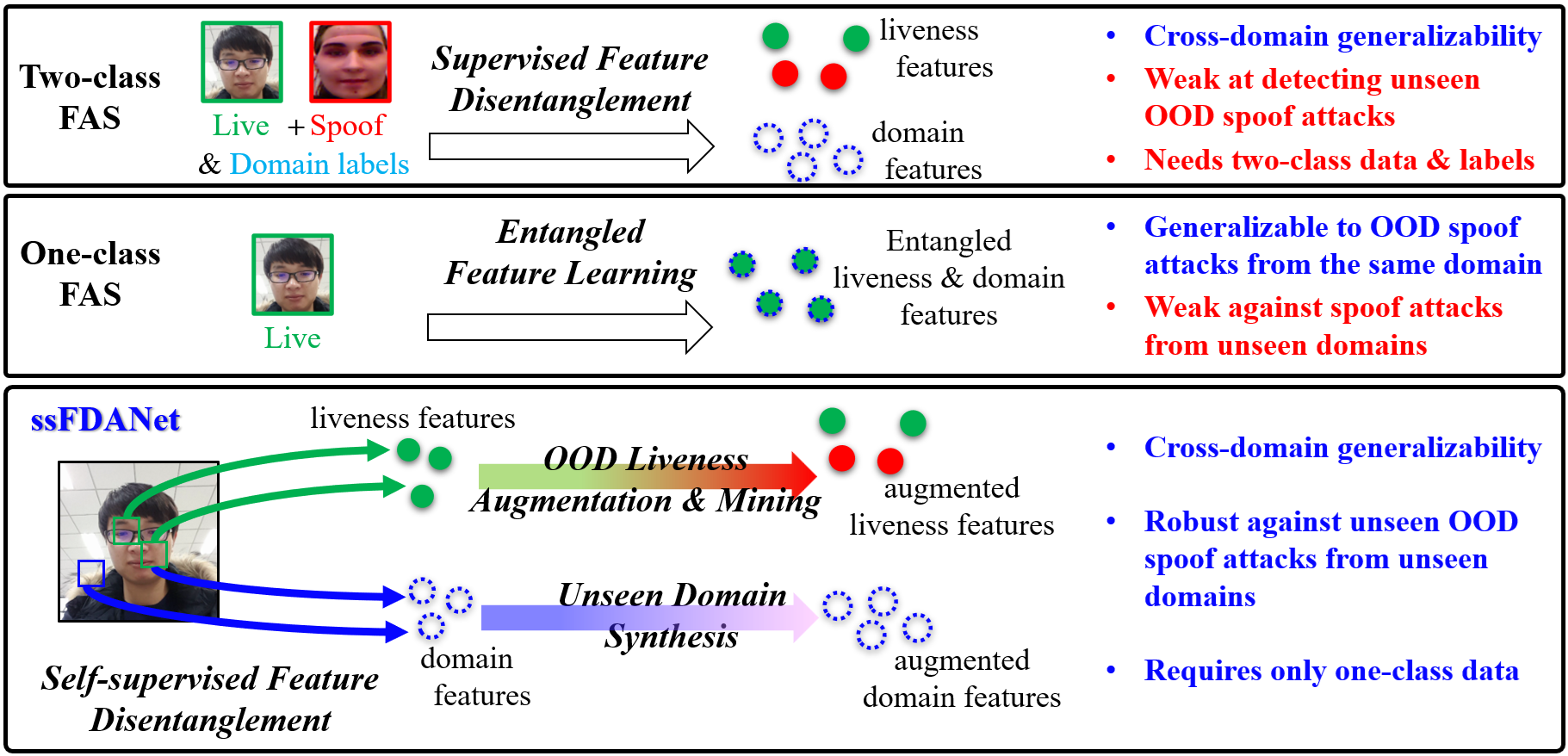} 
    \caption{   
    Differences between conventional face anti-spoofing (FAS) approaches and our one-class FAS method, \textit{i.e.} ssFDANet. Two-class FAS methods rely on supervised feature disentanglement with live/spoof samples and associated binary and domain labels to separate liveness and domain features. In contrast, one-class FAS methods typically learn entangled representations, resulting in limited generalizability and discriminability. ssFDANet introduces a self-supervised, patch-wise feature disentanglement module to explicitly separate liveness and domain features. This enables effective out-of-distribution (OOD) liveness feature augmentation and unseen domain synthesis, allowing ssFDANet to synthesize spoof samples from unseen domains and improve generalization.  
    } 
    \label{fig:idea}  
\end{figure*}

\noindent\textbf{Problem Setting.} \\ 
We consider a one-class FAS setting in which only live face images are available during training, without any spoof images, live/spoof labels, or domain labels. 
Here, \textbf{domain} refers to nuisance factors such as sensors, illumination conditions, capture environments, and background characteristics.
Under this setting, we make two realistic assumptions.
First, the facial foreground and background regions within the same image share similar domain information, as they are captured by the same sensor under identical environmental conditions \cite{huang2025dd,nowara2021benefit}.
Second, different facial patches extracted from a live image share consistent liveness information \cite{huang2025dd,sun2022contrast}.
These assumptions enable self-supervised learning of disentangled liveness and domain representations using live data alone, without external annotations.  
 
The significance of one-class FAS lies in its inherent robustness to unknown spoof attack types. Nevertheless, a major limitation is that the extracted liveness features are often not domain-agnostic, leading to reduced robustness against spoof attacks originating from unseen domains. In contrast, two-class FAS methods---particularly those based on supervised feature disentanglement \cite{wang2022domain,huang2023towards,huang2022learning}---do not suffer from this limitation. Because both genuine and spoof face samples are available in the two-class training set, these methods can utilize supervised disentanglement learning to separate liveness features (used for live/spoof classification) from domain features (reflecting environmental variations), thereby significantly enhancing cross-domain generalizability. However, collecting sufficiently diverse training spoof samples sufficient to cover all potential attack types is impractical in real-world deployments. Therefore, enabling one-class FAS models to inherit the domain generalizability of two-class approaches is a critical yet unresolved challenge for practical face anti-spoofing systems. To achieve this, a fundamental challenge is how to \textbf{disentangle} liveness and domain features from one-class training data in a \textbf{self-supervised} manner. Addressing this challenge would enable downstream feature engineering to enhance both the discriminability and generalizability of one-class FAS models.

To improve one-class FAS, three major challenges must be addressed. First, the absence of spoof samples forces one-class methods to rely on implicit distribution divergence between ``live'' and ``spoof'' data, which limits their ability to learn discriminative features \cite{mcintosh2023inter}.  Second, the visual discrepancy between genuine and spoof faces is often subtle and significantly smaller than that observed in general industrial anomaly detection tasks \cite{tsai2022multi,hyun2023reconpatch}, further exacerbating performance degradation. Third, one-class FAS models typically rely on entangled feature learning, making them vulnerable to domain-specific bias.  More specifically, while two-class FAS methods~\cite{wang2022domain,huang2023towards} achieve better generalizability by leveraging supervised disentanglement to separate liveness and domain features, one-class FAS methods~\cite{huang2021one,lim2020one,huang2024one} cannot leverage such supervision due to the absence of spoof samples and become prone to overfitting to the training domains and sensitive to domain-specific bias. Although recent work \cite{huang2024one} explored pseudo spoof feature generation, the influence of unknown domain variations remains largely unaddressed. 
 
\textcolor{blue}{To tackle these challenges, we propose ssFDANet, a unified framework that disentangles liveness and domain representations while enabling OOD liveness feature augmentation and unseen-domain synthesis under a strict one-class training regime, as illustrated in Fig.~\ref{fig:idea}. }
Based on the assumptions in the problem setting, ssFDANet extracts domain features that remain consistent across regional patches while isolating liveness features using within-image facial patch relationships.
To compensate for the absence of spoof data, ssFDANet further synthesizes pseudo spoof representations through an OOD liveness feature augmentation~\cite{huang2023towards} and a memory-bank-assisted mining strategy, which enables broader latent space exploration and improves coverage of unseen spoof variations.
In addition, to further improve generalization to unseen domains, ssFDANet introduces an adversarial domain learning strategy that augments domain features with plausible real-world characteristics. Together, these components form a unified framework that reduces feature entanglement, alleviates the lack of spoof data, and improves cross-domain robustness for one-class FAS.

\noindent \textbf{Technical Contribution.} \\ 
The main contributions of this work are fourfold.
\textcolor{blue}{First, we propose ssFDANet, a unified framework for one-class face anti-spoofing that jointly integrates self-supervised feature disentanglement, OOD liveness feature augmentation, and reliable domain feature augmentation under a strict one-class training regime.}
Second, we design an OOD liveness feature augmentation and mining mechanism that synthesizes diverse \textcolor{blue}{pseudo-spoof} representations, enhancing the representability and discriminability of liveness features within a one-class framework.
Third, we introduce a reliable domain feature augmentation module that synthesizes unseen domain features with realistic real-world characteristics, improving  the model’s generalizability beyond the training domains.
Finally, extensive experiments on eight public FAS datasets demonstrate that ssFDANet consistently outperforms prior one-class methods and achieves performance comparable to state-of-the-art two-class FAS approaches under both intra-domain and cross-domain protocols.


\section{Related Work} 
\label{sec:Related Work}

\subsection{Two-Class Face Anti-Spoofing}

\subsubsection{Methods involving Supervised Disentangled Feature Learning}

Several two-class FAS methods utilize live/spoof and domain labels to learn disentangled liveness features, aiming to reduce domain interference and effectively distinguish between live and spoof images. Zhang \textit{et al.} \cite{zhang2020face} pioneered the use of supervised disentangled feature learning in intra-domain FAS. Building on this, Wang \textit{et al.} \cite{wang2022domain} proposed approaches to disentangle liveness features from identity or domain cues and recombine them to enhance generalization. However, the limited subject diversity and spoof types in training data often constrain the generalizability of these learned features. 
To mitigate this,  Yue \textit{et al.} proposed CDFTN~\cite{yue2023cyclically}. CDFTN is a domain adaptation framework that synthesizes i) pseudo-labeled samples by recombining domain-invariant liveness features from the source domain and ii) domain-specific content features from the target domain, thereby improving robustness to illumination and spoof-type variations.  
In addition, recent studies have explored supervised disentanglement from different perspectives.
Wang et al. \cite{wang2024tf} introduced language guidance to disentangle content-related features from category-related features for face anti-spoofing.
Li et al. \cite{li2023learning} focused on disentangling spoof traces and identity features across multiple modalities.
Yu et al. \cite{yu2024fourier} further proposed to disentangle domain- and spoof-related features in the Fourier-based frequency space to enhance model generalization.  In our prior work, we extended supervised disentangled feature learning to image- and feature-level augmentation for improved generalization \cite{huang2022learning, huang2023towards}.
However, these methods still require spoof samples and domain annotations, limiting their applicability to one-class FAS. 

\subsubsection{Methods involving Data Augmentation Techniques}

In recent years, various augmentation strategies have been proposed to improve the generalization ability of FAS models and face classification systems. Shao \textit{et al.} developed a domain-transfer augmentation framework that disentangles facial attributes---such as identity, pose/expression, and illumination---to synthesize new face images~\cite{shao2021dotfan}. Huang \textit{et al.} introduced feature mixing to blend domain-specific statistics for domain-invariant learning \cite{huang2022generalized}, while Wang \textit{et al.} further proposed negative data augmentation using color jitter and masking to simulate unseen domains \cite{wang2023domain}. However, spoof samples generated only via such visual transformations provide limited domain variability. To address this limitation, we previously combined data augmentation with supervised disentangled feature learning to enrich spoof diversity and improve cross-domain generalization \cite{huang2022learning, huang2023towards}.
Despite their effectiveness, these approaches still rely on labeled spoof data and are therefore unsuitable for one-class FAS settings.

\subsection{One-Class Face Anti-Spoofing}

One-class FAS methods aim to learn liveness features exclusively from live images to distinguish them from spoof images. Lim \textit{et al.} \cite{lim2020one} and Huang \textit{et al.} \cite{huang2021one} leveraged facial content reconstruction to capture liveness cues. However, in the absence of spoof data, these methods may  learn general facial features rather than discriminative liveness features. Nikisins \textit{et al.} \cite{nikisins2018effectiveness} and Baweja \textit{et al.} \cite{baweja2020anomaly} instead adopted Gaussian Mixture Models (GMMs), modeling the distribution of live features or synthesizing pseudo-spoof features by mixing Gaussian noise with live features.  
To enhance feature discrimination in the absence of spoof cues, we previously introduced a method \cite{huang2024one} that generates pseudo latent spoof features from live images using non-zero spoof cue maps. While this improved training supervision, domain information remained entangled in the learned features, limiting generalizability. 

 In this paper, we address this limitation by explicitly disentangling liveness and domain features to suppress domain-specific bias. The resulting disentangled representations are further used to augment OOD and unseen domain representations, thereby improving both the generalizability and discriminability of one-class FAS models.


\section{Proposed Method} 
\label{sec:Proposed Method}

\begin{figure}[t]
\centering 
\includegraphics[width=1\textwidth]{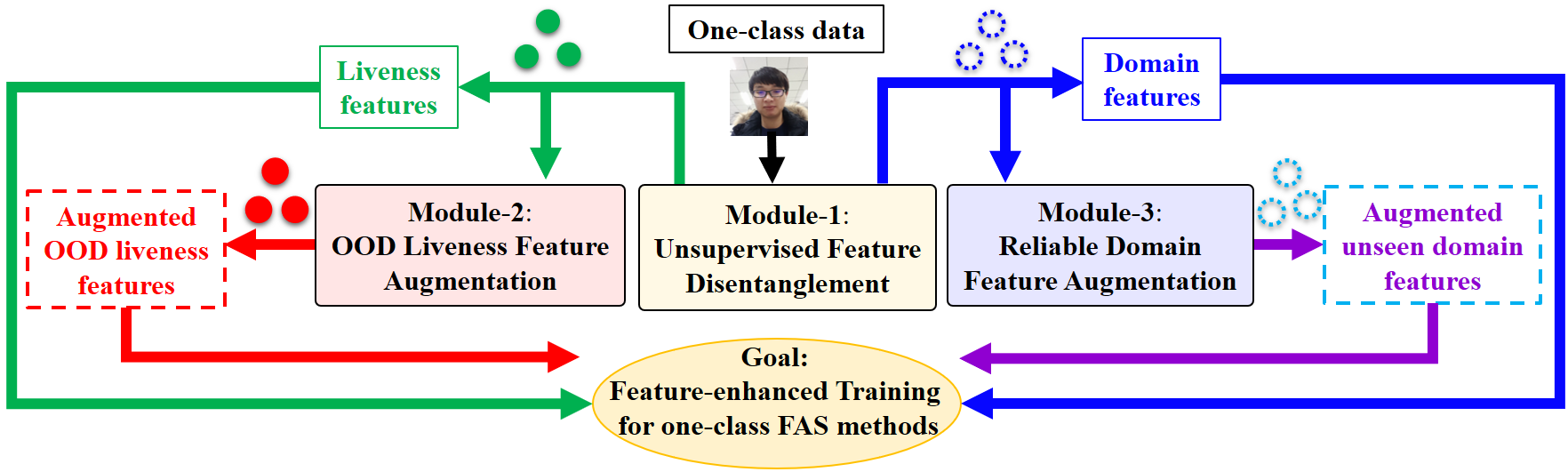} 
\caption{ 
Overview of ssFDANet. ssFDANet consists of three modules: i) a self-supervised feature disentanglement module that separates liveness and domain features, ii) an OOD liveness feature augmentation module that synthesizes OOD spoof classes, and iii) a reliable domain feature augmentation module that generates unseen domain features. These augmented features are used in feature-enhanced training to improve ssFDANet's face anti-spoofing efficacy.  
}  
\label{fig:overview}
\end{figure}

\subsection{Overview}
\label{sec:ssFDANet}
 
\begin{figure*}[t] 
    \centering 
     \includegraphics[width=0.93\textwidth]{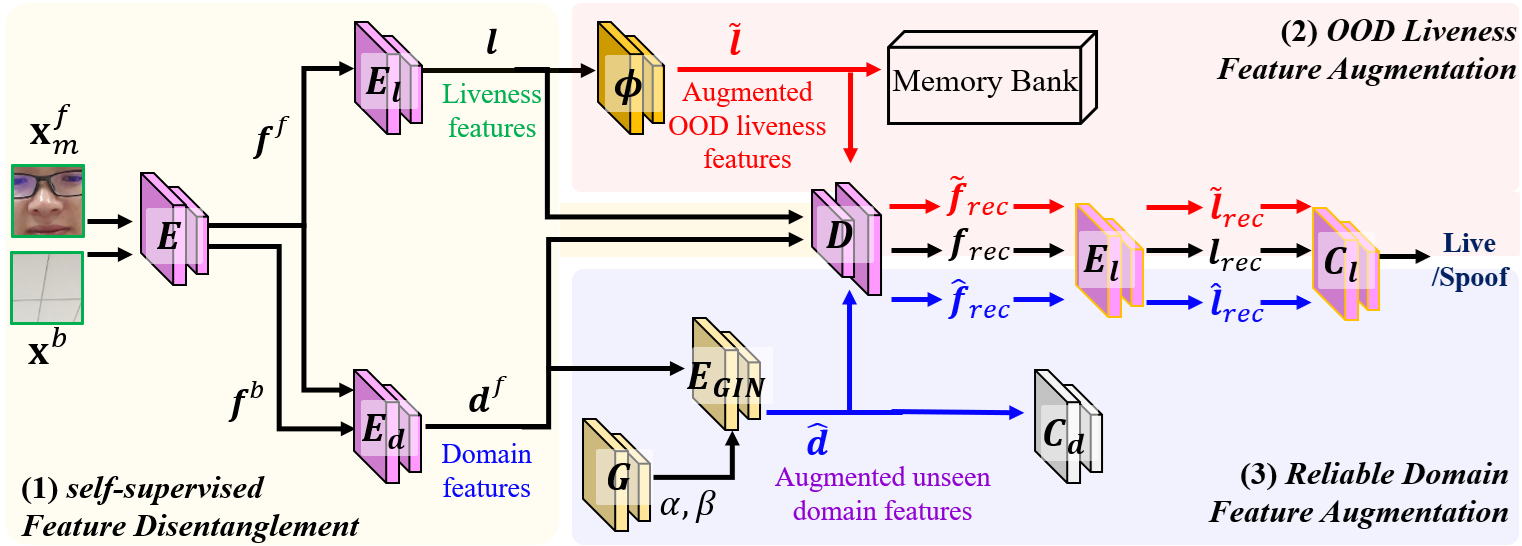} 
    \caption{  
     Flow diagram of ssFDANet. 
     During training, the three modules---\textit{Self-supervised Feature Disentanglement}, \textit{OOD Liveness Feature Augmentation}, and \textit{Reliable Domain Feature Augmentation}---are optimized sequentially within each mini-batch iteration using one-class training data. For deployment, ssFDANet requires only three subnetworks: the general feature encoder $E$, the liveness feature extractor $E_l$, and the liveness classifier $C_l$. The two $E_l$ blocks represent the same subnetwork, and $C_d$ is a frozen pretrained domain classifier. The training pseudo-code is provided in Algorithm~\ref{alg:ssFDANet}.
    }  
    \label{fig:Framework}  
\end{figure*}

\textcolor{blue}{The proposed \textbf{ssFDANet} (Self-supervised Feature Disentanglement and Aug- mentation Network) aims to address three key challenges in one-class face anti-spoofing, namely, (i) feature disentanglement, (ii) spoof data scarcity, and (iii) domain generalization, through three modules: self-supervised feature disentanglement, OOD liveness feature augmentation, and reliable domain feature augmentation.} 
Fig.~\ref{fig:overview} illustrates the overall framework of ssFDANet and the interactions among its three core components. 
The first component is a \textbf{self-supervised feature disentanglement} module that separates liveness and domain information. This addresses a key limitation of conventional one-class FAS methods, which tend to learn entangled and domain-sensitive liveness features, thus improving model robustness. 
The second component is the \textbf{OOD liveness feature augmentation} module. It generates synthetic liveness features for out-of-distribution spoof classes using the liveness representations disentangled from real live samples, thereby supporting spoof detection beyond the training distribution.  
The third component is the \textbf{reliable domain feature augmentation} module, which produces diverse and unseen domain features to improve generalization to novel environments. 
Together, these augmented liveness and domain features enhance the model’s ability to discriminate between live and spoof samples under OOD conditions.
Fig.~\ref{fig:Framework} presents the detailed flow of ssFDANet, and Table\ref{tab:notations} summarizes the used notations.
\textcolor{blue}{Finally, note that ssFDANet's three module was developed
sequentially, with the associated loss terms progressively integrated into the
overall objective. This modular design allows each component and its loss to be validated
individually, ensuring that each loss term serves a specific learning objective.}

\begin{table}[!t]
\centering
\footnotesize
\caption{ Summary of notations} 
\label{tab:notations}
\begin{tabular}{|c|l|}
\hline
    {Symbol} & \multicolumn{1}{c|}{Definition} \\ \hline \hline
    $\mathbf{x}$    &   The live image for training. 
    \\  
    
    $\mathbf{x}^f$    &   The foreground facial part of $\mathbf{x}$. 
    \\ 
    
    $\mathbf{x}^b$    &   The background non-facial part of $\mathbf{x}$. 
    \\ 
  
     
    $\mathbf{x}^f_m$    &   The patch obtained via random masking of $\mathbf{x}^f$.
    \\  \hline

    $E$ & The {general feature encoder}. \\ 
    $E_l$ & The liveness feature extractor. \\ 
    $E_d$ & The domain feature extractor. \\ 
    $D$ & {The general  feature reconstructor.} \\ 
    
    $\phi$  &  {The OOD liveness feature synthesizer.} \\ 

     $E_{GIN}$  & {The GIN encoder for domain feature augmentation.  }
    \\  
     $G$ & The {condition generator} for producing affine \\ 
     & parameters, \textit{i.e.}, $(\alpha, \beta)$, required by $E_{GIN}$. \\ 
     
     $C_l$  &  Binary classifier for liveness vs. non-liveness feature.\\
     $C_d$  & Binary classifier for domain feature vs. non-domain feature.  \\
     \hline
     
     
    $\boldsymbol{f}^{f}$    
        & $\boldsymbol{f}^{f} = E( \mathbf{x}^f_m)$, the {general} feature of $\mathbf{x}^{f}_m$. \\ 
             
    $\boldsymbol{f}^{b}$  & 
    $\boldsymbol{f}^{b} = E(\mathbf{x}^{b})$, the {general} feature of $\mathbf{x}^b$.\\ 
      
    $\boldsymbol{l}$  & $\boldsymbol{l} = E_l(\boldsymbol{f}^{f})$, the liveness feature extracted from $\boldsymbol{f}^{f}$.\\ 
       
    $\boldsymbol{d}^f$  & $\boldsymbol{d}^f = E_{d}(\boldsymbol{f}^{f})$, the domain feature carried by $\boldsymbol{f}^{f}$.\\ 

    $\boldsymbol{d}^b$  & $\boldsymbol{d}^b = E_{d}(\mathbf{f}^{b})$, the domain feature carried by $\boldsymbol{f}^{b}$.\\ 
    
   {$\boldsymbol{f}^f_\mathrm{rec}$}  & The general feature reconstructed by $\boldsymbol{f}^f_\mathrm{rec}=D(\boldsymbol{l}, \boldsymbol{d}^f)$.\\ 

     ${\boldsymbol{l}}_\mathrm{rec}$  & The liveness feature extracted from $\boldsymbol{f}^f_\mathrm{rec}$ by $E_l(\cdot)$.\\

    $\tilde{\boldsymbol{l}}$   & $\tilde{\boldsymbol{l}}=\phi(\boldsymbol{l})$, the augmented OOD liveness feature. \\

    $\tilde{\boldsymbol{l}}_\mathcal{B}$ &  The $\tilde{\boldsymbol{l}}$'s stored in the memory bank $\mathcal{B}$. \\

    $\tilde{\boldsymbol{f}}^f_\mathrm{rec}$
    &  $\tilde{\boldsymbol{f}}^f_\mathrm{rec}=D(\tilde{\boldsymbol{l}}, \boldsymbol{d}^f)$, the general feature reconstructed\\
    & based on the augmented liveness feature $\tilde{\boldsymbol{l}}$.\\

    $\tilde{\boldsymbol{l}}_\mathrm{rec}$ & The liveness feature extracted from $\tilde{\boldsymbol{f}}_\mathrm{rec}$ by $E_l(\cdot)$. \\ 
    
    $\hat{\boldsymbol{d}}$ &     $\hat{\boldsymbol{d}}=E_{GIN}(\boldsymbol{d})$, the augmented domain feature.\\ 
    
    $\hat{\boldsymbol{f}}_\mathrm{rec}$
    & $\hat{\boldsymbol{f}}_\mathrm{rec} = D(\boldsymbol{l}, \hat{\boldsymbol{d}})$, the general feature reconstructed\\
    & based on the augmented domain feature $\hat{\boldsymbol{d}}$.\\ 
 
    $\hat{\boldsymbol{l}}_\mathrm{rec}$  & The liveness feature extracted from $\hat{\boldsymbol{f}}_\mathrm{rec}$ by $E_l(\cdot)$\\ 

    \hline
    
\end{tabular}
\end{table}

\subsection{Module-1: Self-supervised Feature Disentanglement}
\label{sec:Self-supervised-Disentangled Feature Learning}

\begin{figure}[t] 
    \centering  
     \includegraphics[width=0.6\textwidth]{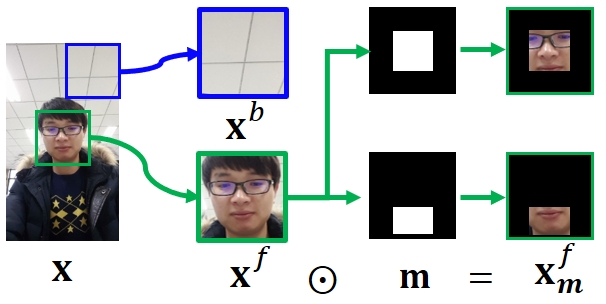}    
    \caption{ 
    Conceptual illustration of patchwise disentanglement of domain and liveness information within a live image. First, the foreground facial part, $\mathbf{x}^{f}$, and the background non-facial part, $\mathbf{x}^{b}$, should share the same domain information  (e.g., sensor, illumination, and capture environment, including background appearance). Second, different masked facial areas, $\mathbf{x}^f_m$, should contain the same liveness information. Consequently, the liveness and domain features can be distilled separately in a patchwise manner. 
    Note that we treat the region detected by 3DDFA \cite{guo2020towards} as the facial region, while the remaining area is regarded as the non-facial area.
   }
    \label{fig:motivation}  
\end{figure}

\begin{figure}[t] 
    \centering 
     \includegraphics[width=0.7\textwidth]{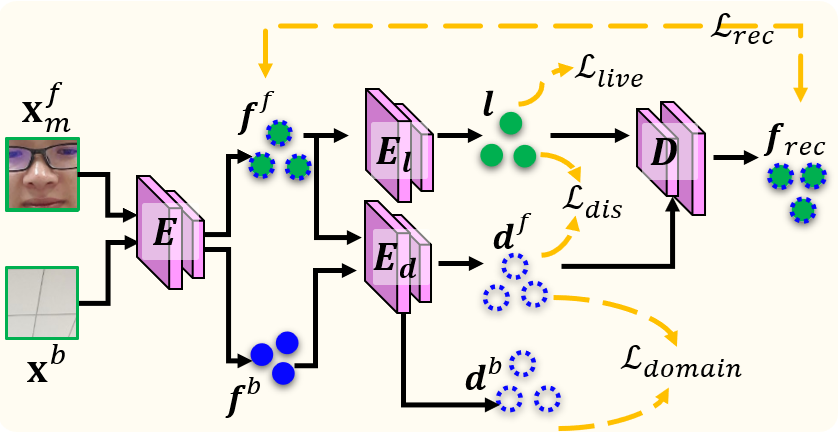}  
    \caption{  Design of the proposed self-supervised feature disentanglement module. It consists of a general feature encoder $E$, a liveness feature extractor $E_{l}$, a domain feature extractor $E_{d}$, and a general feature reconstructor $D$, and it is driven by $\mathcal{L}^\mathrm{ssFD}_{total}=\mathcal{L}_{domain}+\mathcal{L}_{live}+\mathcal{L}_{dis}+\lambda_1\mathcal{L}_{rec}$, described in (\ref{eq:ssfdtotal}). } 
    \label{fig:module1}  
\end{figure}

While two-class FAS methods can learn to extract liveness and domain features separately from both live and spoof samples \cite{huang2022learning,huang2023towards,zhang2020face}, one-class FAS methods cannot, due to the absence of spoof images and corresponding domain labels in their training data. This limitation makes it difficult for one-class approaches to resolve the feature entanglement problem. 
To address this issue, we propose a \textbf{self-supervised feature disentanglement (ssFD)} method that separates liveness and domain features directly from live images, under the one-class FAS setting where spoof data and domain annotations are unavailable. 

The design of the ssFD method is motivated by two key data properties. First, in a live training image, the foreground facial region $\mathbf{x}^{f}$ and the background scene $\mathbf{x}^{b}$ generally share the same domain information, as both are captured by the same camera in a consistent environment~\cite{nowara2021benefit}, as shown in Fig.~\ref{fig:motivation}.  
Second, different patches of the same face $\mathbf{x}^{f}_{m}$ within a video are expected to exhibit similar liveness characteristics~\cite{sun2022contrast}. These two observations motivate our ssFD method to disentangle: 
i) the liveness feature $\boldsymbol{l}$, embedded in the various foreground facial parts $\mathbf{x}^{f}_m$ of the live image $\mathbf{x}$; and  ii) the domain information shared between the foreground facial region $\mathbf{x}^{f}$ and the background region $\mathbf{x}^{b}$. 

To disentangle liveness and domain features, the proposed ssFD method is guided by four loss terms, \textit{i.e.}, i) the domain feature loss $\mathcal{L}_{domain}$, ii) the liveness feature loss $\mathcal{L}_{live}$, iii) the disentanglement loss $\mathcal{L}_{dis}$, and iv) the reconstruction loss $\mathcal{L}_{rec}$.
The domain loss enforces consistency between the domain feature $\boldsymbol{d}^f$ extracted from the foreground patch $\mathbf{x}^f_m$ and the domain feature $\boldsymbol{d}^b$ extracted from the background region $\mathbf{x}^b$, formulated as: 
\begin{equation}
\mathcal{L}_{domain} = \mathbb{E}\{ 1 - \cos( \boldsymbol{d}^f, \boldsymbol{d}^b ) \},
\label{eq: domain extraction loss}
\end{equation}
where $\mathbb{E}{\cdot}$ denotes the expectation, and $\cos(\cdot, \cdot)$ is the cosine similarity. 
Here, $\boldsymbol{d}^f = E_d(E(\mathbf{x}^f_m))$ and $\boldsymbol{d}^b = E_d(E(\mathbf{x}^b))$, where $E(\cdot)$ is a general feature encoder that jointly encodes both liveness and domain information, and $E_d(\cdot)$ is the domain feature extractor. The masked foreground patch $\mathbf{x}^f_m = \mathbf{x}^f \odot \mathbf{m}$ is obtained by applying a binary mask $\mathbf{m}$ to the facial region $\mathbf{x}^f$, as illustrated in Fig.~\ref{fig:motivation}. 

Next, to complement the domain feature loss, we exploit a liveness feature loss $\mathcal{L}_{live}$ to encourage consistency among liveness features extracted from different masked facial regions of the same live image $\mathbf{x}$. Specifically, this loss enforces similarity between liveness features from the $s$-th and $t$-th masked facial patches, that is, 
\begin{equation}
\mathcal{L}_{live} = \mathbb{E}\{1 - \cos(\boldsymbol{l}_{s}, \boldsymbol{l}_{t} )\} \mbox{,}
\label{eq: liveness extraction loss}
\end{equation}
where $\boldsymbol{l}_{s} = E_l(E(\mathbf{x}^f_{m,s}))$ and $\boldsymbol{l}_{t} = E_l(E(\mathbf{x}^f_{m,t}))$ denote the liveness features extracted from the $s$-th and $t$-th masked facial patches of $\mathbf{x}^f$, respectively. Here, $E(\cdot)$ is still the general feature encoder, and $E_l(\cdot)$ is the liveness feature extractor.

Thirdly, the feature disentanglement loss $\mathcal{L}_{dis}$ is used to reduce the correlation between the liveness feature and the domain feature extracted from the same masked facial region   $\mathbf{x}^f_m$. By minimizing the cosine similarity, this loss encourages the two representations to become orthogonal and thereby disentangled. The loss is defined as: 
\begin{equation}
    \mathcal{L}_{dis} = \mathbb{E}\{ \cos( \boldsymbol{l}_{s},  \boldsymbol{d}^f ) \} \mbox{.}
\label{eq: feature disentanglement loss}
\end{equation} 

Finally, to ensure the accuracy of the extracted domain feature $\boldsymbol{d}^f$ and liveness feature $\boldsymbol{l}s$, we define the reconstruction loss $\mathcal{L}_{rec}$ to minimize the discrepancy between (i) the reconstructed global feature $\boldsymbol{f}_{rec} = D(\boldsymbol{l}_s, \boldsymbol{d}^f)$ and (ii) the original global feature of the masked foreground facial image $\boldsymbol{f}^f = E(\mathbf{x}^f_{m})$. This loss is formulated as:
\begin{equation}
\mathcal{L}_{rec} = \mathbb{E}\{ \boldsymbol{f}^f - \boldsymbol{f}_{rec} \} \mbox{,}
\label{eq: reconstruction loss}
\end{equation}
where $D(\cdot, \cdot)$ denotes a learnable general feature reconstructor. This reconstructor is also reused in the later \textbf{OOD liveness feature augmentation} and \textbf{reliable domain feature augmentation} modules.

In summary, the proposed self-supervised feature disentanglement (ssFD) module is trained by minimizing its total loss: 
\begin{equation}
    \mathcal{L}^\mathrm{ssFD}_{total} = \mathcal{L}_{domain} + \mathcal{L}_{live} + \mathcal{L}_{dis} + \lambda_1 \mathcal{L}_{rec} \mbox{,}
    \label{eq:ssfdtotal}
\end{equation}
\noindent 
so that the optimized sub-networks $E$, $E_l$, $E_d$, and $D$, as illustrated in Fig.~\ref{fig:module1}, can be obtained. Note that we set $\lambda_1=10^{-3}$.

\begin{figure}[!t]
\centering
\includegraphics[width=0.65\textwidth]{ 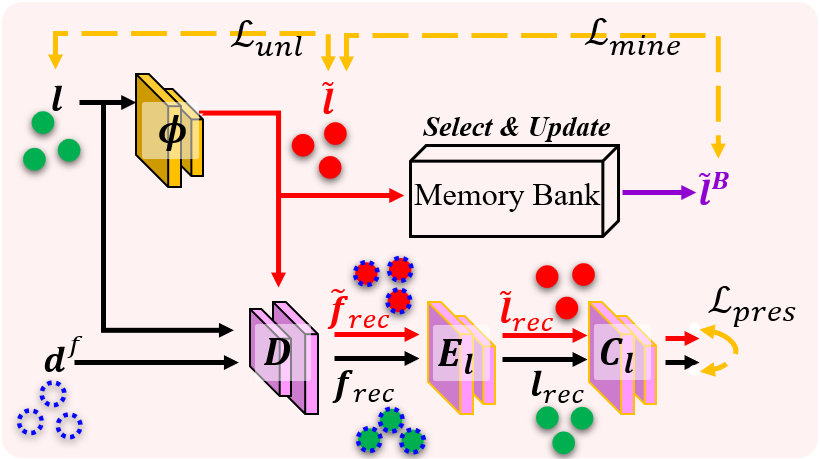}%
\caption{  Design of the proposed out-of-distribution (OOD) liveness feature augmentation module. In addition to the global feature decoder $D$ and the liveness feature extractor $E_{l}$, this module incorporates three additional submodules: an OOD liveness feature adaptor $\phi$, a liveness classifier $C_l$, and a memory bank storing the augmented OOD liveness feature $\tilde{\boldsymbol{l}}_\mathcal{B}$. The training of this module aims to optimize $\phi$ using the unlikeness loss $\mathcal{L}_{unl}$, the liveness preservation loss $\mathcal{L}_{pres}$, and the OOD liveness mining loss $\mathcal{L}_{mine}$, as described in (\ref{eqn:theta_in_OODaugmentation}).}
\label{fig:module2}
\end{figure}

\subsection{Module-2: Out-of-Distribution  Liveness Feature Augmentation}
\label{sec: Liveness Feature Augmentation} 

Because of the absence of spoof face images in the training data, one-class FAS methods typically rely on out-of-distribution (OOD) detection strategies to identify spoofed images.
 Therefore, we propose an OOD liveness feature augmentation strategy to generate pseudo-spoof features and compensate for the absence of spoof data.  
As illustrated in Fig.\ref{fig:module2}, the proposed OOD liveness feature augmentation module comprises five sub-networks, namely,  
i) the OOD liveness feature adaptor $\phi$ (previously introduced in\cite{huang2023towards}), ii) the global feature decoder $D$, iii) the liveness feature extractor $E_{l}$, iv) the liveness classifier $C_l$, and v) a memory bank for storing augmented OOD liveness features $\tilde{\boldsymbol{l}}_\mathcal{B}$.
Among them, the first four form the OOD liveness feature adaptor sub-routine, responsible for synthesizing plausible liveness features for the absent spoof class. The latter two, \textit{i.e.}, $\phi$ and the memory bank, constitute the OOD liveness feature mining sub-routine, which encourages diversity among generated features by exploring underrepresented regions of the latent space. Detailed descriptions of these subroutines are provided below.

\subsubsection{Out-of-Distribution Liveness Feature Adaptor}
\label{sec: AFT Adaptor}

 Given the liveness feature $\boldsymbol{l}_s$ extracted from the $s$-th masked foreground facial patch of $\mathbf{x}^f$, the adaptor $\phi$ synthesizes a plausible OOD liveness feature $\tilde{\boldsymbol{l}}_s$ as:
\begin{equation}
\tilde{\boldsymbol{l}}_s = \phi(\boldsymbol{l}_s) = \mathbf{s} \odot Conv(\boldsymbol{l}_s) + \mathbf{b} \mbox{,}
\label{eq:aft}
\end{equation}
where $\mathbf{s} \sim \mathcal{N}(1, \mathbf{s}_c)$ and $\mathbf{b} \sim \mathcal{N}(0, \mathbf{b}_c)$ are sampled from learnable Gaussian distributions with parameters $\mathbf{s}_c$ and $\mathbf{b}_c$, and $\odot$ denotes element-wise multiplication.
To ensure that the synthesized $\tilde{\boldsymbol{l}}s$ effectively simulates spoof-like representations, we introduce two losses.
The first is the unlikeness loss $\mathcal{L}_{unl}$, which penalizes similarity between the synthetic and input liveness features:
\begin{equation}
\mathcal{L}_{unl} = \mathbb{E}\{ \cos( \tilde{\boldsymbol{l}}_s, \boldsymbol{l}_t ) \} \mbox{.}
\label{eq:dissimilar feature loss}
\end{equation}
The second is the liveness preservation loss $\mathcal{L}_{pres}$, which promotes discriminability while ensuring that feature reconstruction remains faithful:
\begin{equation}
\mathcal{L}_{pres} = -\mathbb{E}\{ \log \big( C_l( \boldsymbol{l}_{rec} ) \big) +\log \big( 1 - C_l( \tilde{\boldsymbol{l}}_{rec} ) \big)\} \mbox{,}
\label{eq:hls_loss}
\end{equation}
where $C_l(\cdot)$ is the classifier distinguishing real liveness features from their augmented counterparts.
Here, $\boldsymbol{l}_{rec}=E_l( D(\boldsymbol{l}, \boldsymbol{d}^f) )$ is the reconstructed liveness feature, and $\tilde{\boldsymbol{l}}_{rec}=E_l( D(\tilde{\boldsymbol{l}}, \boldsymbol{d}^f) )$ is the reconstruction from the augmented feature.
$\mathcal{L}_{pres}$ is minimized when $C_l( \boldsymbol{l}_{rec} )$=1 and $C_l( \tilde{\boldsymbol{l}}_{rec} )$=0, ensuring that
i) $\tilde{\boldsymbol{l}}$ is distinguishable from the real class, and
ii) $D(\cdot, \cdot)$ and $E_l(\cdot)$ preserve feature consistency during reconstruction and extraction.

\subsubsection{Out-of-Distribution Liveness Feature Mining}
\label{sec: Liveness Feature Mining}   
 
The OOD liveness feature mining sub-routine encourages the adaptor $\phi$ to synthesize diverse and unseen OOD liveness features $\tilde{\boldsymbol{l}} = \phi(\boldsymbol{l})$ by exploring the latent feature space for potential spoof attacks.
To this end, we adopt a fixed-size memory bank $\mathcal{B}$   to evaluate how broadly the synthetic features $\tilde{\boldsymbol{l}}$ span the global latent space. Intuitively, the features stored in $\mathcal{B}$ should exhibit minimal mutual dependency, thereby covering a wider spoof-related subspace in the whole latent space.
Based on this principle, for each training batch, an augmented OOD liveness feature $\tilde{\boldsymbol{l}} = \phi(\boldsymbol{l})$ is added to the memory bank $\mathcal{B}$ if its mean absolute cosine similarity with all existing entries in $\mathcal{B}$ is below a predefined threshold $\delta $, i.e., $\mathbb{E}{ |\cos(\tilde{\boldsymbol{l}}, \boldsymbol{l}_\mathcal{B})| } < \delta$, $\forall \boldsymbol{l}_\mathcal{B} \in \mathcal{B}$. This ensures that newly synthesized features are sufficiently distinct from those already stored. If $\mathcal{B}$ is full, it is updated using a first-in-first-out (FIFO) strategy. This mechanism selects a representative and diverse collection of OOD liveness features $\tilde{\boldsymbol{l}}_\mathcal{B}$ that approximates the spoof-related latent subspace.
Accordingly, the OOD liveness feature mining loss $\mathcal{L}_{mine}$ is defined as
\begin{equation}
\mathcal{L}_{mine} = -
\mathbb{E}\{ \log \frac
{\exp(\cos(\tilde{\boldsymbol{l}},\tilde{\boldsymbol{l}}_M))}
{
\exp(\cos(\tilde{\boldsymbol{l}},\tilde{\boldsymbol{l}}_M)) +
\sum_{\forall \tilde{\boldsymbol{l}}_\mathcal{B} } \exp(\cos(\tilde{\boldsymbol{l}},\tilde{\boldsymbol{l}}_\mathcal{B}))} \} \mbox{,}
\label{eq:LS mining}
\end{equation}
where $\tilde{\boldsymbol{l}}_M$ is a randomly masked version of $\tilde{\boldsymbol{l}}$, obtained by element-wise multiplying $\tilde{\boldsymbol{l}}$ with a randomly generated binary mask, 
making $(\tilde{\boldsymbol{l}}, \tilde{\boldsymbol{l}}_M)$ a positive pair. 
In summary, the OOD liveness feature augmentation module is trained using its total loss
\begin{equation}
\mathcal{L}^\mathrm{LiveAug}_{total} =
\mathcal{L}_{unl} +\mathcal{L}_{pres}+ \lambda_2 \mathcal{L}_{mine} \mbox{,}
\label{eqn:theta_in_OODaugmentation}
\end{equation}
\noindent
where $\lambda_2=10^{-1}$. 
This design ensures that: i) the liveness adaptor $\phi$ can generate plausible OOD liveness features that compensate for the absent spoof class; and ii) the synthesized features follow a distribution aligned with that of unseen spoof attacks in the latent space.

\subsection{Module-3: Reliable Domain Feature Augmentation}
\label{sec:Domain Feature Augmentation}
 
Although the previous OOD liveness feature augmentation module can synthesize liveness features $\tilde{\boldsymbol{l}}$ for the pseudo-spoof class, the reconstructed global features $\tilde{\boldsymbol{f}}_{rec}=D(\tilde{\boldsymbol{l}}, \boldsymbol{d}^f)$ still retain the same domain characteristics as the original training data. This limits the generalizability of the model when deployed in unseen domains. To address this, we propose a reliable domain feature augmentation module that synthesizes diverse domain features $\hat{\boldsymbol{d}}$ to simulate real-world variations.  
 
\begin{figure}[!t]
\centering 
\includegraphics[width=0.7\textwidth]{ 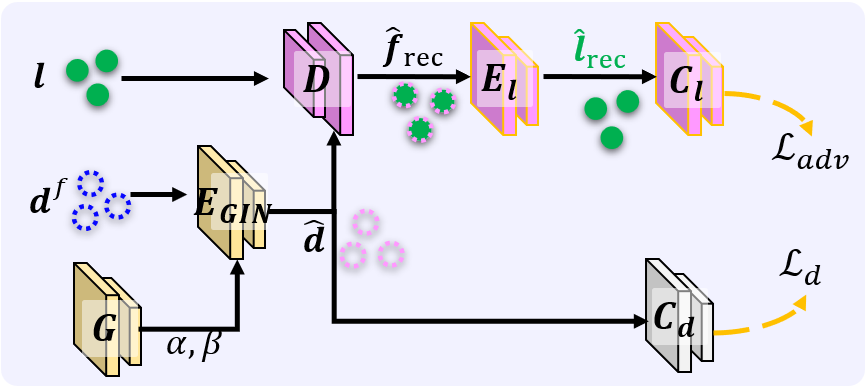} 
\caption{ 
Design of the proposed Domain Feature Augmentation module. Here, $E_{GIN}$, $G$, and $C_d$ denote respectively the Generative Instance Normalization (GIN) encoder, the conditional affine-parameter generator, and a frozen pretrained domain classifier. This module aims to optimize $E_{GIN}$ and $G$ by minimizing $\mathcal{L}^\mathrm{DomainAug}_{total}=\mathcal{L}_{adv}+\mathcal{L}_d$ described in (\ref{eqn:theta_in_G}).  
}  
\label{fig:module3}
\end{figure}

Fig.~\ref{fig:module3} shows the data flow of the reliable domain feature augmentation module. This module consists of two main components: i) the Generative Instance Normalization (GIN) encoder $E_{GIN}$ for synthesizing unseen domain features, and ii) a reliable adversarial domain learning strategy for training the module. 

\subsubsection{Generative Instance Normalization}
\label{sec:Generative Instance Normalization}
 
To synthesize domain features that differ from those in the training set, we design a Generative Instance Normalization (GIN) encoder $E_{GIN}$, which builds upon the adaptive instance normalization strategy \cite{huang2017arbitrary}. GIN consists of a learnable AdaIN module paired with a condition generator $G$ that samples affine parameters $\alpha$ and $\beta$ to transform the domain feature $\boldsymbol{d}^f$:
\begin{equation}
\hat{\boldsymbol{d}} = E_{GIN}\big(\boldsymbol{d}^f | G(\boldsymbol{n}_\alpha, \boldsymbol{n}_\beta)\big) = \alpha \cdot \left(\frac{\boldsymbol{d}^f - \mu_d}{\sigma_d}\right) + \beta \mbox{,}
\label{eq: GIN}
\end{equation}
where $\mu_d$ and $\sigma_d$ denote the mean and standard deviation of $\boldsymbol{d}^f$, respectively. The parameters $\alpha$ and $\beta$ are generated from Gaussian noise vectors $\boldsymbol{n}_\alpha$ and $\boldsymbol{n}_\beta$ via $G(\cdot)$, enabling GIN to produce diverse domain styles instead of fixed shifts.

\subsubsection{Reliable Adversarial Domain Learning}
\label{sec:Reliable ADL}
 
This learning strategy is designed to address two practical concerns. The first is how to synthesize $\hat{\boldsymbol{d}}$ for unseen domains—domains not covered by the training data—while ensuring that these features still capture the underlying characteristics of real-world data. The second is ensuring that the liveness feature extractor $E_l(\cdot)$ remains effective when operating on general features reconstructed using synthetic domain features. 
To this end, we introduce two complementary loss terms. 
 
First, we define an adversarial loss to ensure that the liveness feature extractor $E_l(\cdot)$ can reliably extract spoof-discriminative features from general features constructed with unseen domain inputs. Specifically, given a reconstructed general feature $\hat{\boldsymbol{f}}_{rec} = D(\boldsymbol{l})$, we extract the liveness feature as  $\hat{\boldsymbol{l}}_{rec} = E_l(\hat{\boldsymbol{f}}_{rec})$ and apply the following loss:  
\begin{equation}
\mathcal{L}_{adv} = -\mathbb{E} \{ \log (1 - C_l(\hat{\boldsymbol{l}}_{rec})) \} \mbox{.}
\label{eq:adv_loss}
\end{equation}
This formulation ensures that even when the domain feature  $\hat{\boldsymbol{d}}$ meaningfully shifts the feature distribution, the spoof-discriminative information can still be unaltered during the reconstruction process via $D(\cdot, \cdot)$, faithfully extracted by $E_l$ as the liveness feature $\hat{\boldsymbol{l}}_{rec}$, and correctly classified as spoof ($0$) by the classifier $C_l(\cdot)$. 
Second, to ensure that the synthesized domain features $\hat{\boldsymbol{d}}$ remain consistent with real-world domain characteristics, we introduce a domain entropy loss: 
\begin{equation}
\mathcal{L}_d = -\mathbb{E} \{ \log C_d(\hat{\boldsymbol{d}}) \} \mbox{.}
\label{eq:lsdmloss}
\end{equation}
Here, $C_d(\cdot)$ is a \textit{pretrained} binary classifier  to distinguish domain features (label 1) from liveness features (label 0). This loss encourages $\hat{\boldsymbol{d}}$ to exhibit properties typical of genuine domain signals rather than being confused with liveness features. 
The total training objective for the reliable domain feature augmentation module is defined as:  
\begin{equation}
    \mathcal{L}^\mathrm{DomainAug}_{total} = \mathcal{L}_{adv} + \mathcal{L}_d \mbox{.}
    \label{eqn:theta_in_G}
\end{equation}

\begin{algorithm}[t] 
\caption{Training Procedure of ssFDANet} 
\label{alg:ssFDANet}  
\begin{algorithmic}[1] 
\footnotesize
\REQUIRE ~~\\
	patches $\mathbf{x}^f_m$ and $\mathbf{x}^b$ from a one-class FAS training dataset; maximal training  iteration  $T_{max}$ \\
\ENSURE ~~\\
    The ssFD module consisting of $E$, $E_l$, $E_d$, and $D$; the OOD liveness feature adaptor $\phi$; and, the condition generator $G$, the liveness classifier $C_l$, and the GIN encoder $E_{GIN}$.
    \FOR{training  iteration $t<T_{max}$}
        \STATE
        \textbf{\textit{\# Training Module-1}} \\ 
        \begin{ALC@g}
        \STATE Compute $\mathcal{L}_{domain}$ via (\ref{eq: domain extraction loss}), $\mathcal{L}_{live}$ via (\ref{eq: liveness extraction loss}), $\mathcal{L}_{dis}$ via (\ref{eq: feature disentanglement loss}), and $\mathcal{L}_{rec}$ via (\ref{eq: reconstruction loss}).
        \STATE Update the whole ssFD module, consisting of ${E}$, $E_l$, $E_d$, and $D$, based on $\mathcal{L}^\mathrm{ssFD}_{total}$ in (\ref{eq:ssfdtotal}).
        \end{ALC@g}
       
        \STATE
       \textbf{\textit{\# Training Module-2}}
        \begin{ALC@g}
           \STATE
           Synthesize the OOD liveness feature $\tilde{\boldsymbol{l}}$  via (\ref{eq:aft}).
           \STATE
           Compute $\mathcal{L}_{unl}$ via (\ref{eq:dissimilar feature loss}),  $\mathcal{L}_{pres}$ via (\ref{eq:hls_loss}), $\mathcal{L}_{mine}$ via (\ref{eq:LS mining}).
           \STATE 
           Update $\phi$ based on $\mathcal{L}^\mathrm{LiveAug}_{total}$ in (\ref{eqn:theta_in_OODaugmentation}).
        \end{ALC@g}

        \STATE
       \textbf{\textit{\# Training Module-3}}
        \begin{ALC@g}
            \STATE
            Synthesize unseen domain features $\hat{\boldsymbol{d}}$ via (\ref{eq: GIN}); 
            \STATE 
            Compute $\mathcal{L}_{adv}$ via (\ref{eq:adv_loss}) and  $\mathcal{L}_{d}$ via  (\ref{eq:lsdmloss}).

            \STATE
            Update $G$ and $E_{GIN}$ based on $\mathcal{L}^\mathrm{DomainAug}_{total}$ in (\ref{eqn:theta_in_G}).
        \end{ALC@g}

        \STATE
        \textbf{\textit{\# Feature-enhanced Training:}} 
        \begin{ALC@g}
           \STATE
           Synthesize the OOD liveness features $\tilde{\boldsymbol{l}}$  via (\ref{eq:aft}) and the unseen domain features $\hat{\boldsymbol{d}}$ via (\ref{eq: GIN}).
            \STATE
            Reconstruct $\tilde{\boldsymbol{f}}_{rec}=D(\tilde{\boldsymbol{l}}, \boldsymbol{d}^f)$ and $\hat{\boldsymbol{f}}_{rec}=D(\boldsymbol{l}, \hat{\boldsymbol{d}})$; 
            Compute $\mathcal{L}_{aug}$ via (\ref{eq:rbs_ls_loss}).
            \STATE
            Update $C_l$ and $E_l$ based on $\mathcal{L}_{aug}$.
        \end{ALC@g}
          
    \ENDFOR
\end{algorithmic} 
\end{algorithm}

\subsection{Feature-enhanced Training for ssFDANet}
\label{sec:Augmented Classification} 

Based on the reliable adversarial domain learning strategy described in Sec.~\ref{sec:Reliable ADL}, the representation capability of ssFDANet can be further enhanced by improving its liveness classifier $C_l(\cdot)$ and liveness feature extractor $E_l(\cdot)$. Specifically, $C_l(\cdot)$ and $E_l(\cdot)$ can be refined by minimizing the binary classification loss of $C_l(\cdot)$ on the liveness features extracted by $E_l(\cdot)$, \textit{i.e.},
\begin{equation}
    \mathcal{L}_{aug}
     = -\mathbb{E}\{\log (C_l( E_l(\boldsymbol{f}_{rec})))\}  
  - \mathbb{E}\{ \log (C_{l}( E_l(\hat{\boldsymbol{f}}_{rec })))\}  
  -\mathbb{E}\{ \log (1- C_{l}( E_l(\tilde{\boldsymbol{f}}_{rec})))\}
\mbox{.}
\label{eq:rbs_ls_loss} 
\end{equation}
\noindent
 Here, $\boldsymbol{f}_{rec} = D(\boldsymbol{l}_s, \boldsymbol{d}^f)$ is the reconstructed general feature from the disentangled liveness feature $\boldsymbol{l}_s$ and domain feature $\boldsymbol{d}^f$; $\tilde{\boldsymbol{f}}_{rec}=D(\tilde{\boldsymbol{l}}, \boldsymbol{d}^f)$ is reconstructed using synthetic OOD liveness features for pseudo-spoof faces; and $\hat{\boldsymbol{f}}_{rec}=D(\boldsymbol{l}, \hat{\boldsymbol{d}})$ is reconstructed using the unaltered liveness feature $\boldsymbol{l}$ and the synthesized unseen domain feature $\hat{\boldsymbol{d}}$. This loss is minimized when $C_l( E_l(\boldsymbol{f}_{rec}))=1$ and $C_{l}( E_l(\hat{\boldsymbol{f}}_{rec}))=1$ and $C_{l}( E_l(\tilde{\boldsymbol{f}}_{rec})))=0$.
In summary, this feature-enhanced training step updates both $E_l(\cdot)$ and $C_l(\cdot)$ to improve the liveness discrimination capability of ssFDANet.

\subsection{Training and Inference} 
\paragraph{Training} 
In each training iteration, ssFDANet is optimized in four steps. First, the ssFD module is trained by minimizing the loss in Eq.~(\ref{eq:ssfdtotal}) to update the encoder $E$, the liveness feature extractor $E_l$, the domain feature extractor $E_d$, and the decoder $D$. Second, the OOD liveness feature augmentation module is optimized using Eq.~(\ref{eqn:theta_in_OODaugmentation}) to train the liveness feature adaptor $\phi$. Third, the reliable domain feature augmentation module is trained by minimizing Eq.~(\ref{eqn:theta_in_G}) to learn the condition generator $G$ and the GIN encoder $E_{GIN}$. Finally, the liveness classifier $C_l$ and the liveness feature extractor $E_l$ are further refined using the classification loss in Eq.~(\ref{eq:rbs_ls_loss}). The overall training process of ssFDANet is summarized in Algorithm~\ref{alg:ssFDANet},  where training proceeds in a sequential stage-wise manner at each mini-batch iteration until the maximum number of epochs is reached. 
\paragraph{Testing} 
At inference time, each test image $\mathbf{x}_i$ is classified based on the softmax score $s = C_l(E_l(E(\mathbf{x}_i)))$. Following common practice in previous works \cite{yu2020searching, wang2022domain}, the decision threshold for binary classification is determined using the Youden Index \cite{youden1950index}.

\section{Experiments} 
\label{sec:Experiments}

\subsection{Experimental Settings}
\paragraph{ Implementation Details.} 
 We adopt ResNet-18 as the general feature encoder $E$. The liveness and domain feature extractors, $E_l$ and $E_d$, share the same lightweight architecture implemented as  a lightweight \textit{Conv2D–ReLU} block.  The OOD liveness adaptor $\phi$ is implemented as a $three$-layer convolutional network, while the condition generator $G$ is a $four$-layer MLP. The classifiers $C_l$ and $C_d$  share an identical architecture consisting of a convolutional layer followed by a fully connected layer. 
The decoder $D$ is a single-layer ConvTranspose2d-based network. Except $E$, $E_l$, and $C_l$, all remaining modules are used only during training. We set the memory bank size $\mathcal{B}=10$ and the selection threshold $\delta=0.3$. We train ssFDANet for 50 epochs using MAdam with a learning rate of $5\times10^{-3}$ and a batch size of 4. Additional implementation details are provided in our code repository at \url{https://github.com/jxchong77/ssFDANet}.

\paragraph{Datasets and Evaluation Metrics}
We evaluate the proposed ssFDANet on eight face anti-spoofing datasets, including \textbf{OULU-NPU} \cite{boulkenafet2017oulu} (denoted as \textbf{O}), \textbf{MSU-MFSD} \cite{wen2015face} (denoted as \textbf{M}), \textbf{CASIA-MFSD} \cite{zhang2012face} (denoted as \textbf{C}), \textbf{Idiap Replay-Attack} \cite{chingovska2012effectiveness} (denoted as \textbf{I}), \textbf{SiW} \cite{liu2018learning} (denoted as \textbf{S}), \textbf{3DMAD} \cite{erdogmus2014spoofing} (denoted as \textbf{D}), \textbf{HKBU-MARs} \cite{liu20163d} (denoted as \textbf{H}), and \textbf{CASIA-SURF} \cite{yu2020fas} (denoted as \textbf{U}).
The evaluation metrics include: i) Attack Presentation Classification Error Rate (APCER), ii) Bona Fide Presentation Classification Error Rate (BPCER), iii) Average Classification Error Rate (ACER), iv) Half Total Error Rate (HTER), and v) Area Under the Curve (AUC).

\subsection{Ablation Study}

\begin{table}[t]
\centering
\small 
\setlength{\tabcolsep}{2.5pt}
\caption{
 Ablation study of Module-1, Module-2, and Module-3 on the protocol \textbf{C$\rightarrow$I}  
\label{tab:ablation1}
}
\scalebox{0.85}{ 
\begin{tabular}{|c||c|c|c|c|c|c|c|c|c|c|c|}
\hline 
 & \multicolumn{2}{c|}{\textbf{Module-1}}                                                                                 & \multicolumn{3}{c|}{\textbf{Module-2}}                                                                               & \multicolumn{4}{c|}{\textbf{Module-3}}                                                            & \multicolumn{2}{c|}{\textbf{C$\rightarrow$I}}                  \\ \cline{2-12} 
      & \multicolumn{2}{c|}{$\mathcal{L}^\mathrm{ssFD}_{total}$ }                                                                                   & \multicolumn{3}{c|}{$\mathcal{L}^\mathrm{LiveAug}_{total}$}                                                                                 & \multicolumn{2}{c|}{\textbf{Architectures}}                      & \multicolumn{2}{c|}{$\mathcal{L}^\mathrm{DomainAug}_{total}$} & \multicolumn{1}{c|}{\multirow{2}{*}{HTER}} & \multirow{2}{*}{AUC} 
      
      \\ \cline{2-10}

         &  & $\mathcal{L}_{domain}$ & 
         $\mathcal{L}_{unl}$ &
          &  & 
          & 
          & 
           &
          & 
          & 
         \\ 
         Experiments & $\mathcal{L}_{live}$ &   +$\mathcal{L}_{dis}$+$\mathcal{L}_{rec}$ & 
         + $\mathcal{L}_{pres}$ &
         $\mathcal{L}_{mine}$ & $\mathcal{L}_{cs}$ & 
         AdaIN \cite{huang2023towards}  & 
         $E_{GIN}$ & 
         $\mathcal{L}_{adv}$  &
         $\mathcal{L}_{d}$  & 
          & 
         \\
         \hline
AS-1  & \multicolumn{1}{c|}{\checkmark}   & \multicolumn{1}{c|}{-}                                                                   & \multicolumn{1}{c|}{-}                                                        & \multicolumn{1}{c|}{-}       & -     & \multicolumn{1}{c|}{-}       & \multicolumn{1}{c|}{-}      & \multicolumn{1}{c|}{-}       & -     & \multicolumn{1}{c|}{38.21}                 & 68.67                \\ \hline
AS-2 & \multicolumn{1}{c|}{\checkmark}& \multicolumn{1}{c|}{\checkmark} & \multicolumn{1}{c|}{-}                                                        & \multicolumn{1}{c|}{-}       & -     & \multicolumn{1}{c|}{-}       & \multicolumn{1}{c|}{-}      & \multicolumn{1}{c|}{-}       & -     & \multicolumn{1}{c|}{34.78}                 & 72.45                \\ \hline
AS-3 & \multicolumn{1}{c|}{\checkmark}& \multicolumn{1}{c|}{\checkmark} & \multicolumn{1}{c|}{-}                                                        & \multicolumn{1}{c|}{-}       & \checkmark     & \multicolumn{1}{c|}{-}       & \multicolumn{1}{c|}{-}      & \multicolumn{1}{c|}{-}       & -     & \multicolumn{1}{c|}{33.57}                 & 62.36                \\ \hline
\textcolor{blue}{AS-4} & \multicolumn{1}{c|}{\textcolor{blue}{\checkmark} }& \multicolumn{1}{c|}{\textcolor{blue}{\checkmark} } & \multicolumn{1}{c|}{-}                                                        & \multicolumn{1}{c|}{-}       &  \textcolor{blue}{\checkmark}      & \multicolumn{1}{c|}{-}       & \multicolumn{1}{c|}{ \textcolor{blue}{\checkmark}  }      & \multicolumn{1}{c|}{ \textcolor{blue}{\checkmark} }       &  \textcolor{blue}{\checkmark}     & \multicolumn{1}{c|}{\textcolor{blue}{28.86}}                 & \textcolor{blue}{72.57}                \\ \hline

\textcolor{blue}{AS-5}& \multicolumn{1}{c|}{\checkmark}& \multicolumn{1}{c|}{\checkmark}& \multicolumn{1}{c|}{\checkmark}                                                        & \multicolumn{1}{c|}{-}       & \checkmark     & \multicolumn{1}{c|}{-}       & \multicolumn{1}{c|}{-}      & \multicolumn{1}{c|}{-}       & -     & \multicolumn{1}{c|}{17.07}                 & 88.97                \\ \hline
\textcolor{blue}{AS-6} & \multicolumn{1}{c|}{\checkmark}& \multicolumn{1}{c|}{\checkmark}& \multicolumn{1}{c|}{\checkmark}& \multicolumn{1}{c|}{\checkmark}& \multicolumn{1}{c|}{\checkmark} & \multicolumn{1}{c|}{-}       & \multicolumn{1}{c|}{-}      & \multicolumn{1}{c|}{-}       & -     & \multicolumn{1}{c|}{12.43}                 & 92.49                \\ \hline
\textcolor{blue}{AS-7} & \multicolumn{1}{c|}{\checkmark}& \multicolumn{1}{c|}{\checkmark}& \multicolumn{1}{c|}{\checkmark}& \multicolumn{1}{c|}{\checkmark}& \multicolumn{1}{c|}{\checkmark} & \multicolumn{1}{c|}{\checkmark}       & \multicolumn{1}{c|}{-}      & \multicolumn{1}{c|}{\checkmark}       & -     & \multicolumn{1}{c|}{10.50}                 & 95.56                \\ \hline
\textcolor{blue}{AS-8}& \multicolumn{1}{c|}{\checkmark}& \multicolumn{1}{c|}{\checkmark}& \multicolumn{1}{c|}{\checkmark}& \multicolumn{1}{c|}{\checkmark}& \multicolumn{1}{c|}{\checkmark} & \multicolumn{1}{c|}{-}       & \multicolumn{1}{c|}{\checkmark}      & \multicolumn{1}{c|}{\checkmark}       & -     & \multicolumn{1}{c|}{8.00}                  & 96.22                \\ \hline
\textcolor{blue}{AS-9} & \multicolumn{1}{c|}{\checkmark}& \multicolumn{1}{c|}{\checkmark}& \multicolumn{1}{c|}{\checkmark}& \multicolumn{1}{c|}{\checkmark}& \multicolumn{1}{c|}{\checkmark}& \multicolumn{1}{c|}{-}       & \multicolumn{1}{c|}{\checkmark}      & \multicolumn{1}{c|}{\checkmark}       & \checkmark     & \multicolumn{1}{c|}{\textbf{5.29}}                  & \textbf{98.80 }               \\ \hline

\hline

\end{tabular}
}
\vspace{-0.2cm}
\end{table}

 
Table~\ref{tab:ablation1} demonstrates the contribution of each module in ssFDANet under the \textbf{C$\rightarrow$I} protocol. We first examine \textit{Module-1: self-supervised feature disentanglement}. In AS-1, only $\mathcal{L}_{live}$ is used to aggregate liveness features from live face regions, serving as a baseline. In AS-2, we optimize Module-1 with its full self-supervised loss $\mathcal{L}^{ssFD}_\mathrm{total}$ described in (\ref{eq:ssfdtotal}). Compared to AS-1, AS-2 achieves clear performance gains, indicating that the proposed loss design effectively suppresses domain-related interference in liveness features. Note that in AS-1 and AS-2, since Module-2 is not yet enabled, whether a face image is spoofed is determined by the distance between the liveness feature of the input image and the cluster center of the extracted training liveness features. 
 
Next, \textit{Module-2: OOD liveness feature augmentation} is evaluated. In AS-3, we adopt sampled Gaussian noise as pseudo-spoof features together with the cross-entropy loss $\mathcal{L}_{cs}$ for model training, forming a naive baseline. This strategy performs poorly because Gaussian noise alone cannot adequately approximate the liveness features of the spoof class. 
In \textcolor{blue}{AS-5}, we further introduce $\mathcal{L}_{unl}$ and $\mathcal{L}_{pres}$ to guide the OOD liveness adaptor $\phi$, leading to substantial improvements. Since meaningful liveness representations have already been learned from live images (as shown via AS-2), these losses enable $\phi$ to synthesize features that deviate more effectively from the live-class distribution. By further incorporating the OOD liveness feature mining loss $\mathcal{L}_{mine}$ defined in (\ref{eq:LS mining}), the \textcolor{blue}{AS-6} model further improves performance, confirming that $\mathcal{L}_{mine}$ facilitates broader latent space exploration and more diverse OOD liveness synthesis.  
 
Third, we evaluate \textit{Module-3: reliable domain feature augmentation}. We progressively incorporate components from Module-3 to analyze their effects, as shown in \textcolor{blue}{AS-7, AS-8, and AS-9}. Two observations can be made from these results. 
As for the first observation, under the same adversarial loss $\mathcal{L}_{adv}$, the proposed GIN encoder $E_{GIN}$ \textcolor{blue}{(AS-8)} outperforms the AdaIN-based design \textcolor{blue}{(AS-7)}. This indicates that $E_{GIN}$ is more effective than AdaIN in synthesizing diverse unseen-domain features, as it can sample a broader range of domain transformations conditioned on $G$, whereas AdaIN contains only a limited number of learnable parameters. 
As for the second observation, by introducing the domain entropy loss $\mathcal{L}_{d}$, the performance of \textcolor{blue}{AS-9} further improves. 
\textcolor{blue}{These results suggest the two augmentation branches do not conflict during optimization but instead jointly improve the robustness and generalization of the learned features.}
This indicates that $\mathcal{L}_{d}$ encourages $E_{GIN}$ to synthesize domain features that better match real-world variations, thereby enabling Module-3 to improve domain generalization for one-class FAS.
\textcolor{blue}{In addition, by comparing AS-4 and AS-3, we observe that even when naively sampled Gaussian noise is used as pseudo-spoof features, the synthesized domain features involving real-world variations still help the model better adapt to real-world scenarios, thereby improving performance.}

Finally, owing to the encoder–decoder design adopted for synthesizing liveness features, ssFDANet is also compatible with reconstruction-based anomaly scoring on liveness representations.
All performance results reported in Table~\ref{tab:ablation1}, including AS-8, are obtained using the classifier-based scoring mechanism, i.e., $s = C_l(E_l(E(\mathbf{x})))$. 
When the scoring strategy is switched to reconstruction-based anomaly scoring, defined as $s = \lvert E_l( E(\mathbf{x}) ) - \boldsymbol{f}_{rec} \rvert$, the performance of AS-8 degrades to an HTER of 6.57 and an AUC of 97.99. 
This performance gap indicates that directly classifying disentangled liveness features is more effective than relying on reconstruction residuals. Therefore, ssFDANet adopts classifier-based scoring as the default inference strategy.

\begin{figure}[t]
\centering 
\includegraphics[width=0.8 \textwidth]{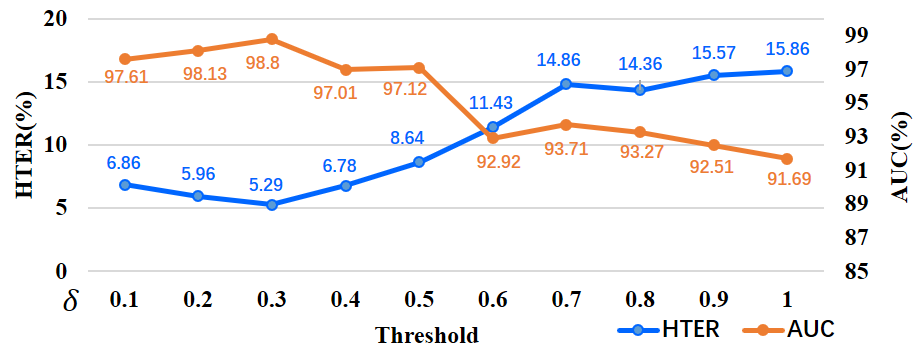} 
\caption{  Sensitivity analysis on the  \textbf{C$\rightarrow$I}  protocol with respect to the selection threshold $\delta$.} 
\label{fig:thresholds}
 \vspace{-0.3cm}
\end{figure}

\subsection{ Sensitivity Analysis on the Selection Threshold $\delta$} 
 Fig.~\ref{fig:thresholds} illustrates the model performance obtained with different selection thresholds $\delta$ for storing augmented OOD liveness features in the memory bank.
Note that when $\delta = 1$, the memory bank update mechanism degenerates into a FIFO strategy without any selection constraint. Without an effective selection strategy, OOD liveness feature mining cannot ensure sufficient diversity in the synthesized features, resulting in degraded performance.
As $\delta$ gradually decreases, the selection strategy encourages augmented OOD liveness features to explore more diverse regions of the latent space, leading to progressively improved performance and achieving the best result at $\delta = 0.3$.
Therefore, $\delta = 0.3$ is selected as the default setting for ssFDANet.

\begin{figure*}
    \centering
    \begin{tabular}{ccc}
    \includegraphics[height=3.6cm]{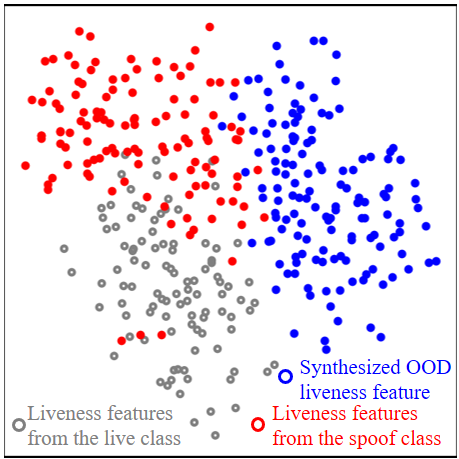} &
    \includegraphics[height=3.6cm]{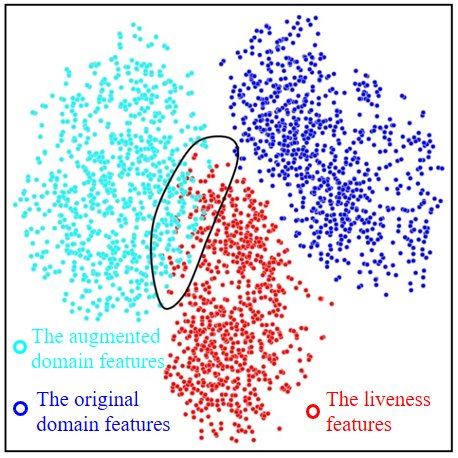} &
    \includegraphics[height=3.6cm]{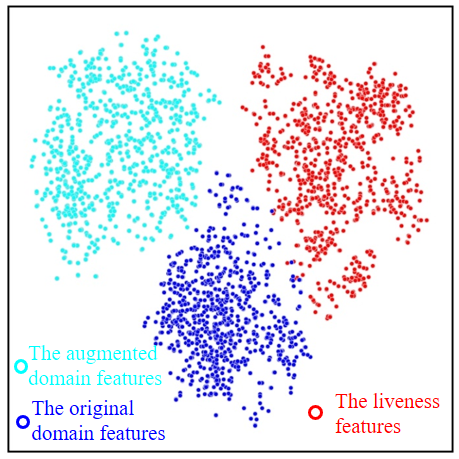} \\
    (a)&(b)&(c)
    \end{tabular}
    \caption{
    $t$-SNE visualizations of the feature distribution on the protocol {C $\rightarrow$ I}. (a) The liveness features. 
    (b) Domain features synthesized by conventional adversarial domain learning employing only $\mathcal{L}_{adv}$.  
    (c)  Domain features synthesized using our reliable adversarial domain learning approach with $\mathcal{L}_{adv}$ + $\mathcal{L}_{d}$.  
    } 
    \label{fig:tsne_all}  
\end{figure*}

\begin{figure}[!t]
\centering
\begin{minipage}{0.48\linewidth}
\centering
\includegraphics[width=\linewidth]{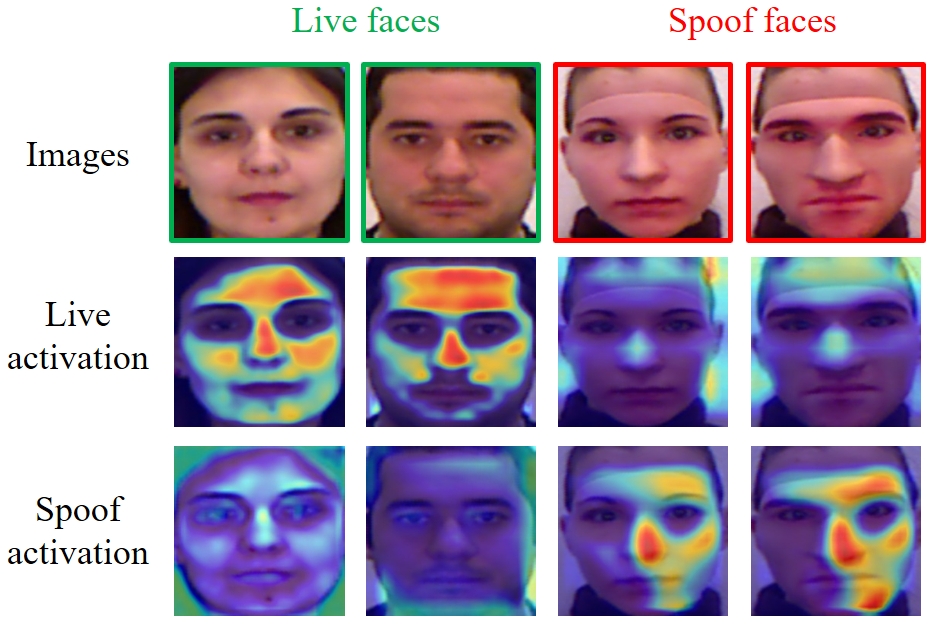} 
\caption{ Live- and spoof-activation maps under the protocol \textbf{[O, S]} $\rightarrow$ \textbf{[D, H, U]}.}
 \label{fig:activation}
\end{minipage}
\hfill
\begin{minipage}{0.48\linewidth}
\centering
\includegraphics[width=\linewidth]{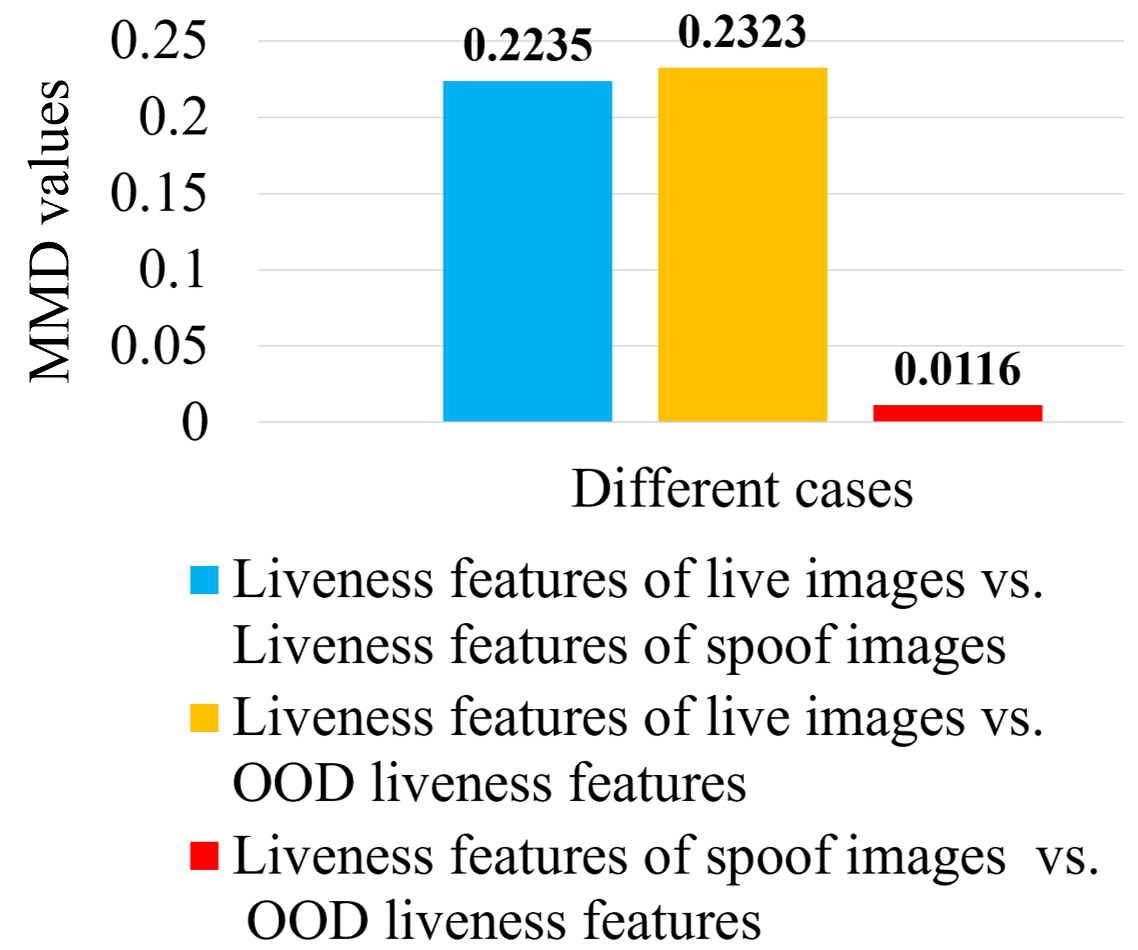} 
\caption{ \textcolor{blue}{MMD distances among different types of liveness features under the protocol \textbf{[C]} $\rightarrow$ \textbf{[I]}.}
}
\label{fig:MMD}
\end{minipage}
 \vspace{-0.3cm}
\end{figure}

\subsection{$t$-SNE Visualization} 

\noindent \textbf{$\bullet$ Synthesized OOD Liveness Features.}
Fig.~\ref{fig:tsne_all}(a) illustrates the distribution of the liveness features on the protocol $\mathbf{C}\rightarrow\mathbf{I}$.  
This figure demonstrates that the synthesized OOD liveness features (blue) not only distribute far away from the live class features (gray), but also scatter widely across regions not covered by the liveness features from the spoof faces (red) in the latent space. This demonstrates that the OOD liveness features synthesized by \textbf{ssFDANet} effectively extend the live training domain, enhancing generalization to unseen attacks.

\noindent  \textbf{$\bullet$ Reliably Augmented Domain Features}: 
Fig.~\ref{fig:tsne_all}(b) and Fig.~\ref{fig:tsne_all}(c) visualize different domain features and the liveness feature under the protocol $\mathbf{C}\rightarrow\mathbf{I}$.
Fig.~\ref{fig:tsne_all}(b)  shows the visualization results from the experiment corresponding to the AS-7 of Table~\ref{tab:ablation1}. 
Since the unseen domain features (cyan) in Fig.~\ref{fig:tsne_all}(b) are synthesized solely under the constraint of $\mathcal{L}_{adv}$, the augmented domain features near the class boundary exhibit overlap with the liveness features (red), as highlighted by the black contour. This overlap  suggests that the augmented domain features may interfere with the discriminability of the liveness features. Conversely, with the incorporation of $\mathcal{L}_d$ (\textit{i.e.,} the AS-8 of Table~\ref{tab:ablation1}), Fig.~\ref{fig:tsne_all}(c) demonstrates that the generated domain features (cyan) are farther from the liveness features (red) and closer to the original domain features (blue) compared to Fig.~\ref{fig:tsne_all}(b). This underscores the efficacy of our proposed reliable domain feature augmentation scheme.


\begin{table*}[!t]
\centering
\footnotesize
\setlength{\tabcolsep}{1pt}
\caption{Comparison of intra-domain face presentation attack detection on \textbf{OULU-NPU}.
The evaluation metrics are APCER(\%) $\downarrow$, BPCER(\%) $\downarrow$, and ACER(\%) $\downarrow$. 
\label{tab:oulu_intra}
}
\scalebox{0.8}{ 
\begin{tabular}{c|c|c|c|c|c|c|c|c|c}
\hline
\textbf{Type} & \textbf{Method} & \textbf{P.} & APCER & BPCER & ACER & \textbf{P.} & APCER & BPCER & ACER\\ 
\hline 

\multirow{5}{*}{\text{2-class}}    & CDCN \cite{yu2020searching} (\textit{CVPR 20}) & \multirow{12}{*}{1} & 0.4 & 1.7  & 1.0 & \multirow{12}{*}{2} & 1.5 & 1.4 & 1.5 \\

                                        &CIFL \cite{chen2021camera} (\textit{TIFS 21}) & & 3.8 & 2.9  & 3.4 & & 3.6 & 1.2  & 2.4  \\      

                                        &PatchNet \cite{wang2022patchnet} (\textit{CVPR 22}) &  & {0.0} & {0.0} & \textcolor{black}{\textbf{0.0}} &  & 0.8 & 1.0 & 0.9 \\

                                        &LDCN \cite{huang2022learnable} (\textit{BMVC 22}) & & 0.0 & 0.0 & \textcolor{black}{\textbf{0.0}} & & 0.8 & 1.0 & 0.9  \\  

                                        &TTN-S \cite{wang2022learning}   (\textit{TIFS 22}) & & 0.4 & 0.0 & 0.2 & & 0.4 & 0.8 & \textcolor{black}{\textbf{0.6}} \\  

                                         & LGON \cite{wang2023learnable}   (\textit{PR 23})  & &  1.5 & 0.0 & 0.8 & & 2.3 & 1.4 & 1.9 \\ 
 \cline{1-2}\cline{4-6} \cline{8-10}

 \multirow{5}{*}{\text{1-class}}   & IQM-GMM \cite{nikisins2018effectiveness}  (\textit{ICB 18})  & &  75.35 & 18.56 & 46.95  & & 41.56 & 27.78 & 34.67 \\
                                & Baweja \textit{et al.} \cite{baweja2020anomaly}  (\textit{IJCB 20})  & & 38.63 & 21.85 & 30.24 & & 51.81 & 19.83 & 35.82 \\
                                & Lim \textit{et al.} \cite{lim2020one} (\textit{Access 20}) & & 43.54 & 36.5 & 40.02 & & 72.19 & 18.5 & 45.35  \\
                                & AAE \cite{huang2021one}  (\textit{CCBR 21})  & &  47.13 & 26.67 & 36.9  & &  37.28 & 39.0 & 38.14\\
                                & OC-SCMNet \cite{huang2024one}  (\textit{CVPR 24})  & &  20.83 & 26.15 & 23.49  & &  22.05 & 28.81 & 25.43\\
 \cline{2-2}\cline{4-6} \cline{8-10}
                                &  \textbf{ssFDANet}  (Ours) & & 16.38 & 29.17 & \textcolor{black}{\textbf{22.77}}  & & 24.72 & 26.06 & \textcolor{black}{\textbf{25.39}}\\	
                                
\cline{1-2}\cline{4-6} \cline{8-10}
 
\hline

\multirow{5}{*}{\text{2-class}}    & CDCN \cite{yu2020searching} (\textit{CVPR 20}) & \multirow{12}{*}{3} & 2.4$\pm$1.3 & 2.2$\pm$2.0 & 2.3$\pm$1.4 & \multirow{12}{*}{4} &  4.6$\pm$4.6 & 9.2$\pm$8.0 & 6.9$\pm$2.9  \\

                                        &CIFL \cite{chen2021camera} (\textit{TIFS 21}) & & 3.8$\pm$1.3 & 1.1$\pm$1.1 & 2.5$\pm$0.8 & & 5.9$\pm$3.3 & 6.3$\pm$4.7 & 6.1$\pm$4.1   \\      

                                        &PatchNet \cite{wang2022patchnet} (\textit{CVPR 22}) & &  1.8$\pm$1.47 & 0.56$\pm$1.24 & 1.18$\pm$1.26 & &  2.5$\pm$3.81 & 3.33$\pm$3.73 & 2.90$\pm$3.00   \\

                                        &LDCN \cite{huang2022learnable} (\textit{BMVC 22})& & 4.55$\pm$4.55 & 0.58$\pm$0.91 & 2.57$\pm$2.67 & & 4.50$\pm$1.48 & 3.17$\pm$3.49 & 3.83$\pm$2.12   \\      

                                        &TTN-S \cite{wang2022learning}   (\textit{TIFS 22}) & &  1.0$\pm$1.1 & 0.8$\pm$1.3 & \textcolor{black}{\textbf{0.9$\pm$0.7}} & & 3.3$\pm$2.8 & 2.5$\pm$2.0 & \textcolor{black}{\textbf{2.9$\pm$1.4}}   \\      

&  LGON \cite{wang2023learnable}   (\textit{PR 23})  & &  1.3$\pm$0.9 & 0.8$\pm$0.9 & 1.0$\pm$0.6  & & 3.3$\pm$2.6& 3.3$\pm$4.1 & 3.3$\pm$2.6 \\ 
                                        
 \cline{1-2}\cline{4-6} \cline{8-10}

 \multirow{5}{*}{\text{1-class}}   & IQM-GMM \cite{nikisins2018effectiveness}  (\textit{ICB 18})  & &  57.17$\pm$16.79 & 16.5$\pm$6.95 & 36.83$\pm$5.35 & & 53.42$\pm$14.08 & 16.67$\pm$8.38 & 35.04$\pm$3.95   \\      
                                & Baweja \textit{et al.} \cite{baweja2020anomaly}  (\textit{IJCB 20})  & &  45.39$\pm$12.82 & 18.28$\pm$16.21 & 31.83$\pm$6.99 &  & 60.25$\pm$16.49 & 10.67$\pm$10.37 & 35.46$\pm$5.43   \\      
                                & Lim \textit{et al.} \cite{lim2020one} (\textit{Access 20}) & &  38.51$\pm$13.08 & 39.52$\pm$11.13 & 39.02$\pm$2.16 & & 36.91$\pm$10.24 & 20.5$\pm$8.01 & 28.07$\pm$5.32   \\      
                                & AAE \cite{huang2021one}  (\textit{CCBR 21})  & &  26.62$\pm$13.67 & 52.93$\pm$16.09 & 39.77$\pm$3.74 & & 26.33$\pm$18.5 & 40.17$\pm$29.04 & 33.12$\pm$8.9   \\      
                                & OC-SCMNet \cite{huang2024one}  (\textit{CVPR 24})  & &  27.10$\pm$12.57 & 20.55$\pm$11.12 & 23.83$\pm$3.14 & & 16.41$\pm$14.00 & 11.66$\pm$9.42 & 14.04$\pm$4.90   \\ 
                                \cline{2-2}\cline{4-6} \cline{8-10}
                                &  \textbf{ssFDANet}  (Ours) & &  19.04$\pm$8.94 & 21.33$\pm$7.09 & \textcolor{black}{\textbf{20.19$\pm$3.47}} & & 4.08$\pm$3.29 & 6.83$\pm$9.06 & \textcolor{black}{\textbf{5.46$\pm$5.30}}   \\      
\cline{1-2}\cline{4-6} \cline{8-10}
  
\hline

\end{tabular}}
 \vspace{-0.3cm}
\end{table*}

\begin{table}[t]
\centering
\scriptsize 
\setlength{\tabcolsep}{1pt}
\caption{Comparison results for cross-domain testing on \textbf{{[}M, I{]} $\rightarrow$ C} and \textbf{{[}M, I{]} $\rightarrow$ O}.    
\label{tab:limited_cross_type}}
\begin{tabular}{c|c|c|c|c|c}
\hline
\multirow{2}{*}{\textbf{Type}} & \multirow{2}{*}{\textbf{Method}} &  \multicolumn{2}{c|}{\textbf{{[}M,I{]} $\rightarrow$ C}} & \multicolumn{2}{c}{\textbf{{[}M,I{]} $\rightarrow$ O}} \\ \cline{3-6} 
 & & \multicolumn{1}{c|}{HTER $\downarrow$} & AUC  $\uparrow$& \multicolumn{1}{c|}{HTER $\downarrow$} & AUC  $\uparrow$ \\ \hline 
\multirow{6}{*}{\shortstack{ 2-class}}  &SSAN-M \cite{wang2022domain}  (\textit{CVPR 22}) & \multicolumn{1}{c|}{30.00} & 76.20 & \multicolumn{1}{c|}{29.44} & 76.62 \\  
&LDCN \cite{huang2022learnable}  (\textit{BMVC 22}) & \multicolumn{1}{c|}{22.22} & 82.87 & \multicolumn{1}{c|}{\textcolor{black}{21.54}} & \textcolor{black}{86.06} \\  
&CADMA \cite{huang2022generalized}  (\textit{ICASSP 22}) & \multicolumn{1}{c|}{28.75} & 75.78 & \multicolumn{1}{c|}{\textcolor{black}{26.28}} & \textcolor{black}{80.46} \\  
&DiVT-M \cite{liao2023domain}  (\textit{WACV 23}) & \multicolumn{1}{c|}{\textcolor{black}{\textbf{20.11}}} & \textcolor{black}{\textbf{86.71}}  & \multicolumn{1}{c|}{23.61 } & 85.73 \\ 
&NDA-FAS \cite{wang2023domain}  (\textit{TIFS 23}) & \multicolumn{1}{c|}{\textcolor{black}{21.78}} & \textcolor{black}{83.01}  & \multicolumn{1}{c|}{21.41} & {86.35} \\ 
&DFANet \cite{huang2023towards}  (\textit{ICME 23}) & \multicolumn{1}{c|}{\textcolor{black}{20.67}} & \textcolor{black}{84.87}  & \multicolumn{1}{c|}{\textbf{18.61}} & \textbf{89.52} \\ 

 & HPDR \cite{hu2024rethinking} (\textit{CVPR 24}) & \multicolumn{1}{c|}{22.22} & 85.54 & \multicolumn{1}{c|}{21.07} & 87.53 \\
\hline 
\multirow{5}{*}{\shortstack{ 1-class}} & IQM-GMM \cite{nikisins2018effectiveness}  (\textit{ICB 18}) & 45.81 & 39.74 & 35.0 & 37.01\\		 
 & Baweja \textit{et al.} \cite{baweja2020anomaly}  (\textit{IJCB 20}) & 27.33 & 78.50 & 32.01 & 72.19 \\
& Lim \textit{et al.} \cite{lim2020one} (\textit{Access 20}) & 43.56 & 53.6 & 39.19 & 64.11 \\
& AAE \cite{huang2021one}  (\textit{CCBR 21}) & 46.67 & 47.28 & 48.52 & 47.99\\	& OC-SCMNet \cite{huang2024one}  (\textit{CVPR 24}) & 21.67 & \textbf{85.30} & 22.03 & \textbf{84.28}\\		 

 \cline{2-6} 

 & \textbf{ssFDANet}  (Ours)  & \textcolor{black}{\textbf{21.66}}  & \textcolor{black}{82.42} & \textcolor{black}{\textbf{21.92}} & 82.84 \\   

 & [95 \% Confidence Interval]  &
 \begin{tabular}[c]{@{}c@{}}   [19.7791,    \\  24.1409]  \end{tabular}
 
& \begin{tabular}[c]{@{}c@{}}   [78.47,   \\  84.23]   \end{tabular}  &  \begin{tabular}[c]{@{}c@{}}  [20.0080,  \\  24.1520]    \end{tabular}    & 

\begin{tabular}[c]{@{}c@{}}  [79.32, \\ 84.78]    \end{tabular} 
 
 \\ 
 \hline

\end{tabular}
 \vspace{-0.3cm}
\end{table}

\subsection{Activation Visualization} 
 
Figure~\ref{fig:activation} presents the activation maps generated under the cross-domain protocol \textbf{[O, S]} $\rightarrow$ \textbf{[D, H, U]}, which includes 3D mask attacks. For live samples (green boxes), the live activation maps show strong responses across facial regions, including the forehead, while the spoof activation maps display minimal activation, indicating that the model correctly suppresses spoof-related responses in live faces.  
In contrast, for spoof samples involving 3D masks (red boxes), the spoof activation maps are concentrated on the mask-covered facial areas, with noticeably lower responses in the uncovered forehead region. Conversely, the live activation maps of these spoof images exhibit higher responses in the forehead area, suggesting that the model recognizes inconsistencies between covered and uncovered regions. 
These results demonstrate that ssFDANet effectively localizes meaningful facial regions and captures spoof-specific cues, enabling robust discrimination between live and spoof faces, even in 
cross-domain scenarios. 

\subsection{\textcolor{blue}{Measuring Feature Differences Across Different Liveness Features}} 

\textcolor{blue}{
In Figure~\ref{fig:MMD}, we use Maximum Mean Discrepancy (MMD) to measure the feature differences among three types of liveness features: (1) those extracted from live images, (2) those from spoof images, and (3) the generated OOD liveness features. The results show that the generated OOD liveness features exhibit clear separation from the liveness features of live images in terms of MMD, while remaining much closer to the liveness features of spoof images. This observation suggests that the generated OOD liveness features are not trivial outliers but approximate plausible spoof representations, thereby providing useful supervision for one-class training and improving robustness to unseen attack types. 
}



\begin{table}[t]
\scriptsize
\caption{
Comparison results for cross-domain testing on \textbf{C $\rightarrow$ I} and \textbf{I $\rightarrow$ C}. }
\centering
\begin{tabular}{|c |c |c | c | c | c | c |}
 \hline   
    \textbf{Type}   & \textbf{Method}   
                    & \textbf{C $\rightarrow$ I}   
                    & \textbf{I $\rightarrow$ C } 
                    &  \# param.  
                    &  FLOPs  
                    &  FPS   \\
       &    
                    & HTER  $\downarrow$  
                    & HTER  $\downarrow$
                    & (training)  
                    & (training)  
                    & (testing) 
\\   
\hline           
    & STASN \cite{yang2019face} (CVPR 19) & 31.5 &  30.9 & - & - & - \\  
    & CDCN \cite{yu2020searching} (CVPR 20) & 15.5 & 32.6 &  2.25M  &  47.52G  &  292 \\ 
2-class
    & CIFL \cite{chen2021camera} (TIFS 21) & 17.6 & - & - & - & - \\ 
    & PatchNet \cite{wang2022patchnet} (\textit{CVPR 22}) & \textbf{9.9}&  \textbf{26.2} & - & - & - \\ 
    & NDA-FAS \cite{wang2023domain} (\textit{TIFS 23}) &   12.5 & 31.2 &  17.44M  &  96.13G & 110 \\ 
 \hline
    & IQM-GMM \cite{nikisins2018effectiveness}  (\textit{ICB 18}) & 31.93 & 48.44 &  159.74M  & 15.49G  &  31  \\		 
    & Baweja \textit{et al.} \cite{baweja2020anomaly}  (\textit{IJCB 20}) & 46.29 & 29.44 &  145.03M   &  15.48G & 210  \\
 1-class 
    & Lim \textit{et al.} \cite{lim2020one} (\textit{Access 20}) & 37.36 & 39.78 & 8.86M  &  20.62G & 388  \\ 
    & AAE \cite{huang2021one} (\textit{CCBR 21}) & 20.0 & 26.9 &  2.42M  & \ 5.72G &  816 \\
    & OC-SCMNet \cite{huang2024one} (\textit{CVPR 24}) & 7.29 & 17.44 &  5.92M   & 47.08G & 373 \\
    \cline{2-7} 
 
    &  \textbf{ssFDANet}  (Ours) &  \textbf{5.29} & \textbf{17.43} &  \multirow{2}{*}{6.54M}   & \multirow{2}{*}{2.98G}      & \multirow{2}{*}{875}   \\
    &   [95\% Confidence Interval]   &  [4.86, 5.76]  &   [16.14, 20.60]  &  &  & \\  
\hline   
\end{tabular}
    \label{tab:cross_testing_CI}
    \vspace{-0.3cm}
\end{table}

\begin{table}[t]
\scriptsize
\setlength{\tabcolsep}{2pt}
\caption{ Comparison of cross-domain testing on \textbf{{[}O, S{]} $\rightarrow$ \textbf{[}D, H, U{]}}. 
 }
\centering
\begin{tabular}{c|c|c|c|c|c|c|c}
\hline 
\multirow{2}{*}{\textbf{Type}} & \multirow{2}{*}{\textbf{Method}} &  \multicolumn{2}{c|}{ \textbf{[O,S] $\rightarrow$ D }} & \multicolumn{2}{c|}{ \textbf{[O,S] $\rightarrow$ H }} & \multicolumn{2}{c}{ \textbf{[O,S] $\rightarrow$ U }} \\  \cline{3-8} 
 & & \multicolumn{1}{c|}{HTER  $\downarrow$} & {AUC $\uparrow$ } & \multicolumn{1}{c|}{HTER  $\downarrow$} & AUC $\uparrow$ & \multicolumn{1}{c|}{HTER $\downarrow$} & AUC $\uparrow$ \\ \hline
2-  &  Auxiliary \cite{liu2018learning}  
    & 0.29 & 99.04  &  14.64 &  88.32  &  37.28 &  53.14 \\  
class &NAS \cite{yu2020fas} 
    & \textbf{0.22} & 99.31 & 15.13 & 88.91 & 37.68 & \textbf{72.83} \\   
 & LDCN \cite{huang2022learnable} 
    & 1.49 & \textbf{99.91} & \textbf{8.75} & \textbf{95.60} & \textbf{33.54}&  60.44 \\
\hline 

 & IQM-GMM \cite{nikisins2018effectiveness} 
    & 43.83 & 43.43 & 19.14 & 80.53 & 38.18 & 66.18\\  
 & Baweja \textit{et al.} \cite{baweja2020anomaly}  
    & 37.86 & 45.80 & 35.65 & 68.90 & 41.74 & 49.85  \\  
1- & Lim \textit{et al.} \cite{lim2020one}  
    & 27.69 & 75.47 & 35.19& 62.98 & 37.34 & 64.24 \\  
class & AAE \cite{huang2021one} 
    & 22.48& 78.62 & 31.22 & 73.77& 45.24 & 53.48 \\ 
& OC-SCMNet \cite{huang2024one} 
    & 1.47 & 99.87 & 7.08 & 86.84 & \textbf{10.61} & \textbf{90.75} \\  \cline{2-8} 
 
 &\textbf{ssFDANet}  (Ours)& \textbf{1.17} & \textbf{99.92} &\textbf{5.55} & \textbf{96.65} & 13.76 & 92.60 \\ 
  &   [95\% Confidence Interval] &  \begin{tabular}[c]{@{}c@{}}   [1.16,  \\ 1.21]    \end{tabular}  & 
  \begin{tabular}[c]{@{}c@{}}   [99.63,  \\  99.99]    \end{tabular}  &
  \begin{tabular}[c]{@{}c@{}}   [5.23,  \\  6.15]   \end{tabular} & 
   \begin{tabular}[c]{@{}c@{}}  [93.08,  \\  99.61]   \end{tabular}  & 
   \begin{tabular}[c]{@{}c@{}}   [11.75, \\ 15.23]    \end{tabular}
  & \begin{tabular}[c]{@{}c@{}}   [86.72, \\ 96.42]    \end{tabular} \\
\hline   
 
\end{tabular}
\label{tab:cross_domain_3dmask}
\end{table}

\subsection{Intra-Domain Testing}
Table~\ref{tab:oulu_intra} demonstrates that the proposed \textbf{ssFDANet}, as a one-class approach, achieves a significant advantage over all other one-class FAS methods in the intra-domain testing scenario on \textbf{OULU-NPU}~\cite{boulkenafet2017oulu}. In this setting, because the spoof attacks in the training and testing data share highly similar characteristics, two-class FAS methods generally outperform one-class methods in distinguishing live from spoof samples. Nevertheless, despite being a one-class method, \textbf{ssFDANet} substantially narrows the performance gap with two-class methods, highlighting the effectiveness of its self-supervised feature disentanglement and OOD liveness feature augmentation modules.

\subsection{Cross-Domain Testing} 


Table \ref{tab:limited_cross_type} shows the cross-domain testing results for the protocols \textbf{{[}M, I{]} $\rightarrow$ C} and \textbf{{[}M, I{]} $\rightarrow$ O}, designed to detect print and replay attacks based on the settings used in \cite{shao2019multi}. In this setting, datasets \textbf{M} and \textbf{I} are used for model training because of their substantial domain variation, while datasets \textbf{C} and \textbf{O} are used for testing.
In Table \ref{tab:limited_cross_type}, \textbf{ssFDANet} outperforms all one-class FAS methods in terms of HTER while achieving competitive performance compared to two-class FAS methods. This result implies that two-class FAS methods may overfit the training data, leading to performance degradation when there are cross-domain shifts in spoofing characteristics. In contrast, the proposed \textbf{ssFDANet} can operate without prior knowledge of the spoof class, demonstrating superior domain generalizability in detecting unseen spoof attacks in cross-domain scenarios.

Following the experimental configurations in \cite{liu2018learning}, we conducted cross-domain evaluations using the datasets \textbf{C} and \textbf{I}, as shown in Table \ref{tab:cross_testing_CI}. While dataset \textbf{I} consists solely of low-resolution data, dataset \textbf{C} contains both high- and low-resolution data. Consequently, the protocol \textbf{I $\rightarrow$ C} is inherently more challenging than \textbf{C $\rightarrow$ I}, because models trained exclusively on low-resolution data often struggle to generalize to high-resolution data, which presents a persistent challenge for two-class FAS methods.
In contrast, as a one-class FAS method, \textbf{ssFDANet} significantly improves detection performance and even outperforms two-class FAS methods in this scenario. These results highlight the effectiveness of \textbf{ssFDANet} in cross-domain testing.

Finally, we followed the experimental settings used in \cite{yu2020fas} to perform another cross-domain evaluation on 3D-mask attack types, with results detailed in Table \ref{tab:cross_domain_3dmask}. Our model was trained on live images from datasets \textbf{O} and \textbf{S} and tested on datasets \textbf{D}, \textbf{H}, and \textbf{U}. 
Table \ref{tab:cross_domain_3dmask} shows that two-class FAS methods, typically trained on live images and spoof images with print and replay attacks, often struggle to generalize to previously unseen attacks such as 3D-mask attacks, as observed in the \textbf{{[}O, S{]} $\rightarrow$ \textbf{U}} protocol. In contrast, \textbf{ssFDANet} outperforms all FAS methods, including both one-class and two-class approaches, on the \textbf{{[}O, S{]} $\rightarrow$ \textbf{U}} and \textbf{{[}O, S{]} $\rightarrow$ \textbf{H}} protocols. 
Additionally, \textbf{ssFDANet} achieves competitive performance against two-class methods on the \textbf{{[}O, S{]} $\rightarrow$ \textbf{D}} protocol. These results further demonstrate the efficacy of the proposed \textbf{ssFDANet} in handling cross-domain scenarios.

\begin{table}[!t]
\scriptsize
\setlength{\tabcolsep}{2pt}
\caption{ 
\textcolor{blue}{Disentanglement consistency evaluation on \textbf{{[}O, S{]} $\rightarrow$ \textbf{[}D, H, U{]}}. }
 }
\centering
\color{blue}
\begin{tabular}{ c|c|c|c|c|c|c}
\hline 
  \multirow{2}{*}{\textbf{Method}} &  \multicolumn{2}{c|}{ \textbf{[O,S] $\rightarrow$ D }} & \multicolumn{2}{c|}{ \textbf{[O,S] $\rightarrow$ H }} & \multicolumn{2}{c}{ \textbf{[O,S] $\rightarrow$ U }} \\  \cline{2-7} 
  & \multicolumn{1}{c|}{HTER  $\downarrow$} & {AUC $\uparrow$ } & \multicolumn{1}{c|}{HTER  $\downarrow$} & AUC $\uparrow$ & \multicolumn{1}{c|}{HTER $\downarrow$} & AUC $\uparrow$ \\ \hline
  
 \textbf{ssFDANet} (Original) &  1.17  & 99.92  & 5.55 &  96.65 & 13.76 & 92.60 \\ 
 \textbf{ssFDANet} (Augmented)& 1.19 & 99.51 & 6.81 & 98.17 & 15.82 & 91.09 \\ 
  
\hline   
 
\end{tabular}
\vspace{-0.3cm}
\label{tab:Disentanglement_Consistency}
\end{table}

\subsection{\textcolor{blue}{Disentanglement Consistency Evaluation}}

\textcolor{blue}{
To evaluate the consistency of the disentangled representations, Table~\ref{tab:Disentanglement_Consistency} compares the model performance on the original testing data and randomly augmented data. Specifically, we first disentangle the liveness and domain features from the original testing samples and then use the general feature reconstructor $D$ to randomly recombine these features to synthesize augmented general features. The trained model is then applied to re-disentangle the augmented data, and classification is performed using the resulting liveness features. The model achieves highly consistent performance on both the original and reconstructed data, suggesting that the learned representations maintain a consistent functional separation between liveness-related and domain-related information rather than degenerating into arbitrary partitions of the feature space. 
}

\subsection{ Discussion}

\noindent  \textbf{$\bullet$ Computational Efficiency.}  
In Table~\ref{tab:cross_testing_CI}, we report the computational statistics of different FAS methods.
ssFDANet adopts a ResNet-18 backbone as the general feature encoder $E$ and introduces additional modules during training (\textit{e.g.}, $E_l$, $E_d$, $E_{GIN}$, $\phi$, $G$, $C_d$, $C_l$), resulting in a total parameter size of 6.54M.
At test time, inference only involves the feature encoder and the liveness classifier (\textit{i.e.}, $s = C_l(E_l(E(\mathbf{x}_i)))$), while all training-specific modules are disabled.
Consequently, ssFDANet incurs lower memory usage during inference and achieves higher FPS.

\noindent \textbf{$
\bullet$ Training Stability.} 
 To assess training stability, we conduct five independent runs and report the 95\% confidence intervals in Tables~\ref{tab:limited_cross_type}, \ref{tab:cross_testing_CI}, and \ref{tab:cross_domain_3dmask}. On relatively easier protocols (\textit{e.g.}, \textbf{[O,S] $\rightarrow$ D} and \textbf{[O,S] $\rightarrow$ H}), ssFDANet exhibits narrower confidence intervals, indicating consistent performance across different runs. On more challenging protocols (\textit{e.g.}, \textbf{[O,S] $\rightarrow$ U}), the confidence intervals become moderately wider, reflecting increased variability under larger domain shifts.
Overall, ssFDANet demonstrates stable training behavior across different evaluation protocols.

\noindent  \textbf{$
\bullet$ Limitation.}  
This work focuses on image-level face anti-spoofing, following the setting adopted by most existing FAS studies. As a result, temporal cues such as motion patterns or temporal moire artifacts are not explored. In addition, the proposed self-supervised feature disentanglement may be less effective under challenging conditions, including dynamic backgrounds, camera-induced motion blur, and partial facial occlusions (\textit{e.g.}, masks).

\noindent \textbf{$
\bullet$ Future Improvements.}  
Several directions may be explored to further extend this work.
First, expanding the ``prompt vocabulary'' of liveness cues~\cite{liu2024cfpl} could enhance self-supervised feature disentanglement learning.
Second, although ssFDANet adopts a ResNet-based backbone, alternative architectures such as Mixture-of-Experts~\cite{liu2024moeit} may improve feature representation and robustness to unseen target domain.
Third, more diverse generative and domain-shift strategies~\cite{liu2021face} could be incorporated to further strengthen OOD liveness feature augmentation.

\section{Conclusion} 
\label{sec:Conclusion}

\textcolor{blue}{
In this paper, we introduced a Self-supervised Feature Disentanglement and Augmentation Network (\textbf{ssFDANet}) to address the challenges of one-class Face Anti-Spoofing (FAS). The proposed framework disentangles liveness and domain features from live images through a self-supervised mechanism, and further enhances model robustness through OOD liveness feature augmentation and reliable domain feature augmentation. 
The contribution of this work lies in organizing these components into a unified framework that enables disentanglement, pseudo-spoof synthesis, and domain generalization to be learned jointly under a strict one-class training regime. 
Extensive experiments demonstrate that \textbf{ssFDANet} significantly outperforms previous one-class FAS methods and achieves competitive performance compared with state-of-the-art two-class FAS models. These results validate the effectiveness of the proposed framework in improving the representability, discriminability, and generalization capability of one-class FAS systems.
}
\bibliographystyle{elsarticle-num-names} 
\bibliography{egbib}

\end{document}

%% file: Titlepage.tex
\begin{frontmatter}

\title{
Self-supervised Feature Disentanglement and Augmentation Network for One-class Face Anti-spoofing
\tnoteref{t1}} 

 \author[1,2]{Pei-Kai Huang\fnref{fn1}}
\ead{alwayswithme@gapp.nthu.edu.tw}

\author[2]{Jun-Xiong Chong \fnref{fn1}}
\ead{jxchong@gapp.nthu.edu.tw}

\author[2]{Ming-Tsung Hsu}
\ead{xmc510063@gapp.nthu.edu.tw}

 \author[2]{Fang-Yu Hsu}
\ead{shellyhsu@gapp.nthu.edu.tw}

\author[2]{Yi-Ting Lin}
\ead{yitinglin@gapp.nthu.edu.tw}

\author[2]{Jun-Ren Chen}
\ead{jr.chen.088@gapp.nthu.edu.tw}

\author[2]{Kai-Heng Chien}
\ead{ss113062582@gapp.nthu.edu.tw}

\author[3]{Hao-Chiang Shao \corref{cor1}}
\ead{shao.haochiang@gmail.com}

\author[2]{Chiou-Ting Hsu}
\ead{cthsu@cs.nthu.edu.tw}

\cortext[cor1]{The corresponding author is Prof. Hao-Chiang Shao.}
\fntext[fn1]{These two authors contributed equally}

\affiliation[1]{organization={College of Computer and Cyber Security, Fujian Normal University},
                city={Fuzhou},
                country={China}}

\affiliation[2]{organization={Department of Computer Science, 
                              National Tsing Hua University},
                city={Hsinchu},
                country={Taiwan}}

\affiliation[3]{organization={Institute of Data Science and Information Computing, 
                             National Chung Hsing University}, 
                city={Taichung}, 
                country={Taiwan} }
\begin{abstract}
Face anti-spoofing (FAS) aims to protect face recognition systems by detecting and rejecting fraudulent presentation attacks. FAS methods are commonly categorized into two paradigms: one-class and two-class approaches. While two-class FAS methods often achieve higher accuracy on known attack types, they tend to generalize poorly in open-set scenarios. In contrast, one-class methods offer better robustness against out-of-distribution (OOD) spoof attacks but typically suffer from inferior overall accuracy. To address this trade-off, we propose a novel one-class FAS framework, named  \textbf{Self-supervised Feature Disentanglement and Augmentation Network (ssFDANet)}, designed to incorporate advantageous characteristics from both paradigms. Specifically, we design a self-supervised feature disentanglement module  that separates domain-related and liveness-related features using only live face data. To further enhance performance, we incorporate two complementary feature augmentation modules: one for synthesizing diverse liveness features representative of unseen spoof types, and the other for generating domain features beyond the training distribution to improve domain robustness. Extensive experiments demonstrate that the proposed \textbf{ssFDANet} significantly improves robustness against unseen spoof attacks and domain shifts, outperforming prior one-class FAS methods and achieving performance comparable to state-of-the-art two-class approaches. 

\end{abstract}
 
\begin{keyword}
Face anti-spoofing \sep one-class classification \sep  disentangled feature learning \sep   affine feature transformation \sep  adversarial feature learning
\end{keyword} 
\end{frontmatter}